\documentclass[10pt,twocolumn,letterpaper]{article}

\usepackage{iccv}
\usepackage{times}
\usepackage{epsfig}

\usepackage{graphicx}
\usepackage{amsmath,amssymb} 
\usepackage{color}

\usepackage{times}
\usepackage{epsfig}
\usepackage{algorithm}
\usepackage{subfigure}
\usepackage{caption}	
\usepackage{booktabs}
\usepackage{url}
\usepackage{float}
\usepackage{placeins}
\usepackage{flushend}

\usepackage{mathtools}
\DeclareMathOperator*{\argmin}{argmin}
\usepackage[breaklinks=true,bookmarks=false]{hyperref}

\iccvfinalcopy 

\def\iccvPaperID{10} 

\ificcvfinal\fi
\begin{document}

\title{Distributed Bundle Adjustment}

\author{Karthikeyan Natesan Ramamurthy\\
IBM Research\\
{\tt\small knatesa@us.ibm.com}
\and
Chung-Ching Lin \\
IBM Research\\
{\tt\small cclin@us.ibm.com}
\and
Aleksandr Aravkin\\
University of Washington\\
{\tt\small saravkin@uw.edu}
\and
Sharath Pankanti\\
IBM Research\\
{\tt\small sharat@us.ibm.com}
\and
Raphael Viguier\\
University of Missouri-Columbia\\
{\tt\small rvbb3@missouri.edu}
}

\maketitle
\ificcvfinal\thispagestyle{empty}\fi

\begin{abstract}
Most methods for Bundle Adjustment (BA) in computer vision are either centralized or operate incrementally. This leads to poor scaling and affects the quality of solution as the number of images grows in large scale structure from motion (SfM). Furthermore, they cannot be used in scenarios where image acquisition and processing must be distributed. We address this problem  with a new distributed BA algorithm. Our distributed formulation uses alternating direction method of multipliers (ADMM), and, since each processor sees only a small portion of the data, we show that robust formulations improve performance. We analyze convergence of the proposed algorithm, and  illustrate numerical performance, accuracy of the parameter estimates, and scalability of the distributed implementation in the context of synthetic 3D datasets with known camera position and orientation ground truth. The results are comparable to an alternate state-of-the-art centralized bundle adjustment algorithm on synthetic and real 3D reconstruction problems. The runtime of our implementation scales linearly with the number of observed points. 

\end{abstract}

\section{Introduction}
\label{sec:intro}
Estimating accurate poses of cameras and locations of 3D scene points from a collection of images obtained by the cameras is a classic problem in computer vision, referred to as \textit{structure from motion} (SfM). Optimizing for the camera parameters and scene points using the corresponding points in images, known as {\it Bundle Adjustment} (BA), is an important component of SfM~\cite{hartley2003multiple,heinly2015reconstructing,triggs2000bundle}. 


Many recent approaches for BA can be divided into three categories: (a) those that pose BA as non-linear least squares \cite{konolige2010sparse,lourakis2009sba,triggs2000bundle}, (b) those that decouple the problem in each camera using a triangulation-resection procedure for estimation \cite{mitra2008scalable,pritt2014fast}, and (c) those that pose and solve BA in a linear algebraic formulation \cite{fusiello2015solving}. Some important considerations of these methods are reducing the computational complexity by exploiting the structure of the problem  \cite{agarwal2010bundle,byrod2010conjugate,lourakis2009sba}, incorporating robustness to outlier observations or correspondence mismatches \cite{aravkin2012student,zhang2006robust}, distributing the computations or making the algorithm incremental \cite{indelman2012incremental,kopf2014first,wu2013towards,wu2011multicore, erikssonconsensus} and making the algorithm insensitive to initial conditions \cite{fusiello2015solving}. In this paper, we develop robust distributed BA over camera and scene points. 

Our approach is ideally suited for applications where image acquisition and processing must be distributed, such as in a network of unmanned aerial vehicles (UAVs). We assume that each UAV in the network has a camera and a processor; each camera acquires an image of the 3D scene, and the processors in the different UAVs cooperatively estimate the 3D point cloud from the images. Therefore, we use the terms camera, processor, and UAV in an equivalent sense throughout the paper. We also assume that corresponding points from the images are available (possibly estimated using a different distributed algorithm), and are only concerned about estimating the 3D scene points given the correspondences.

Robust approaches, such as \cite{aravkin2012student,zhang2006robust}, are typically used to protect world point and camera parameter estimates from effects of {\it outliers}, which for BA are incorrect point correspondences that have gone undetected. In contrast, we use robust formulations to {\it accelerate consensus} in the distributed formulation. Depending on how distribution is achieved, every processor performing computation may see {\it only a small portion} of the total data, and attempt to use it to infer its local parameters. Small sample means can be extreme, even when the original sample is well-behaved (i.e. even when re-projection errors are truly Gaussian). In the limiting case, each processor may only base its computation on {\it one data point}, and therefore outliers are guaranteed to occur (from the point of view of individual processors) as an artifact of distributing the computation. Hence we hypothesize that using robust losses for penalizing re-projection errors, and quadratic losses for enforcing consensus improves performance. 


\begin{figure}[tbh]
\begin{center}
\subfigure[]{
\includegraphics[width=4.0cm]{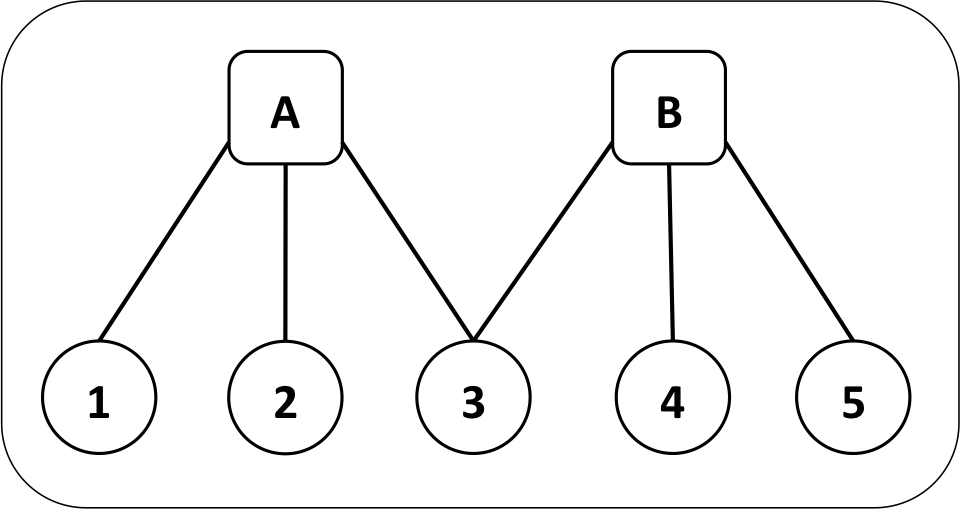}
}
\subfigure[]{
\includegraphics[width=4.5cm]{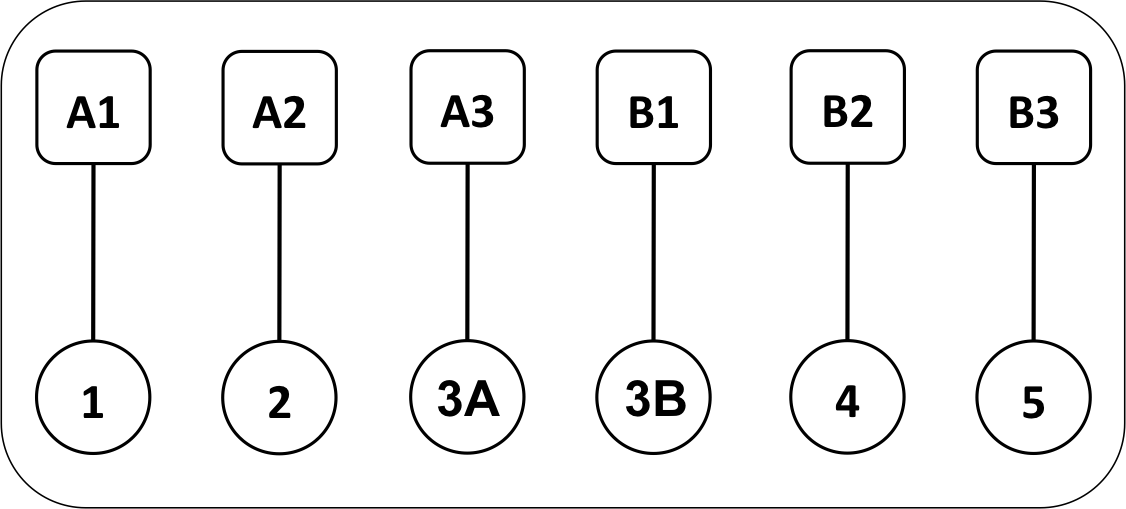}
}
\end{center}


\caption{(a) Original configuration of cameras A, B and scene points $1, 2, 3, 4, 5$, (b) distributing both the camera parameter and scene point estimation with the constraints A1 = A2 = A3, B1 = B2 = B3, and 3A = 3B.}
\label{fig:WP_CP_dist}
\end{figure}

Our proposed robust BA approach supports a natural distributed parallel implementation. We distribute the world points and camera parameters as illustrated for a simple case of $2$ cameras and $5$ scene points in Figure \ref{fig:WP_CP_dist}. The algorithm is developed using distributed alternating direction method of multipliers (D-ADMM)~\cite{boyd2011distributed}. 
Each processor updates its copy of a set of parameters, while the updated estimates and dual variables ensure consensus. 
Distributing {\it both} the world points and the camera parameters yields iterations with $O(l)$ required operations in a serial setting, where $l$ is the total number of 2D observations. 
In a fully parallel setting, it is possible to bring the time complexities down to $O(1)$ per iteration, 
a vast improvement compared to traditional and sparse versions of BA, whose complexities are $O((m+n)^3)$ and $O(m^3+mn)$ respectively \cite{lourakis2009sba} 
(with $m$ and $n$ the number of cameras and 3D scene points). 
We also exploit the sparsity of the camera network, since not all cameras observe all scene points.

Another optimization-based distributed approach for BA was recently proposed \cite{erikssonconsensus} \footnote{The initial version of our method was proposed at the same time as \cite{erikssonconsensus}}. Authors of~\cite{erikssonconsensus} distributed camera parameters, and performed synthetic experiments using an existing 3D point cloud reconstruction, perturbing it using moderate noise, and generating image points using known camera models. 
We go further, distributing both world points and camera parameters in a flexible manner, and we implement the entire BA pipeline for 3D reconstruction: 
performing feature detection, matching corresponding points, and applying the robust distributed D-ADMM BA technique in real data settings. 

The rest of the paper is organized as follows. We provide background in Section \ref{sec:background}, and present the new formulation in Section \ref{sec:approach}. We show experiments and results on synthetic and real data in Section \ref{sec:expt}, and conclude with Section \ref{sec:concl}.

\section{Background}
\label{sec:background}

\subsection{The camera imaging process}
\label{sec:cam_im}
We denote the $m$ camera parameter vectors by $\{y_j\}_{j=1}^m$ , the $n$ 3D scene points as $\{x_i\}_{i=1}^n$, and the 2D image points as $\{z_{ij}\}$. Each 2D image point $z_{ij} \in \mathbb{R}^2$ is obtained by the transformation and projection of a 3D scene point $x_i \in \mathbb{R}^q$ by the camera $y_j \in \mathbb{R}^p$.  BA is an inverse problem, where  camera parameters and  3D world points are estimated from the observations $\{z_{ij}\}$.
The forward model is a non-linear camera transformation function  $f(x_i,y_j)$. 

The number of image points is typically much smaller than $mn$, since not all cameras image all scene points. 
The camera parameter vector ($y_j$) usually includes  position, Euler angles, and focal length. 
In this discussion, we assume focal length is known for simplicity, and $y_j \in \mathbb{R}^6$ comprises Euler angles $\alpha, \beta, \gamma$ and the translation vector $t \in \mathbb{R}^3$. 

Denote the diagonal focal length matrix as $K \in \mathbb{R}^{3 \times 3}$, with the first two diagonal elements set to the focal length and the last element set to $1$. The rotation matrix is represented as $R = R_3(\gamma) R_2(\beta) R_1(\alpha)$, where $R_1,R_2,R_3$ are rotations along the three axes of $\mathbb{R}^3$. The camera transformation is now given as $\tilde{z} = Rx+t$. The final 2D image point $z$ is obtained by a perspective projection, with co-ordinates given by
\begin{equation}
\label{eqn:persp_proj}
z_1 = \frac{\tilde{z}_1}{\tilde{z}_3}, z_2 = \frac{\tilde{z}_2}{\tilde{z}_3}.
\end{equation} 

\subsection{Bundle adjustment}
\label{sec:BA}
Given the 2D points in multiple images that represent the same scene point, BA is typically formulated as a nonlinear least squares problem: 
\begin{equation}
\label{eqn:BA}
\min_{\{x_i\},\{y_j\}} \sum_{j = 1}^m \sum_{i \in S(j)} \|z_{i,j} -f(x_i,y_j)\|_2^2.
\end{equation} The set $S(j)$ contains $i$ if the scene point $i$ is imaged by the camera $j$. The number of unknowns in this objective is $3n+6m$, and hence it is necessary to have at least this many observations to obtain a good solution;
in practice the number of observations is much larger. Problem~\eqref{eqn:BA} is solved iteratively, 
with descent direction ($\delta_x, \delta_y$) found by replacing $f$ in~\eqref{eqn:BA} by its linearization 
\[
f(x + \delta_x,y+\delta_y) \approx f(x,y) + J(x) \delta_x + J(y) \delta_y,
\] 
where $J(x) = \partial_x f, J(y) = \partial_y f$. The Levenberg-Marquardt (LM) algorithm~\cite{more1978levenberg} is often used for BA. 


The naive LM algorithm requires $O((m+n)^3)$ operations for each iteration, and memory on the order of $O(mn(m+n))$,
since we must invert of an $O(m+n) \times O(m+n)$ matrix at each iteration. 
However, exploiting matrix structure and using the Schur complement approach proposed in \cite{lourakis2009sba}, 
the number of arithmetic operations can be reduced to $O(m^3+mn)$, and memory use to $O(mn)$. 
Further reduction can  be achieved by exploiting secondary sparse  structure \cite{konolige2010sparse}.
The conjugate gradient approaches~in \cite{agarwal2010bundle,byrod2010conjugate} 
can reduce the time complexity to $O(m)$ per iteration, making it essentially linear in the number of cameras.

Another popular approach to reduce the computational complexity involves decoupling of the optimization by explicitly estimating the scene point using back-projection in the \textit{intersection} step and estimating the camera parameters in the \textit{resection} step \cite{pritt2014fast}. 
The resection step decouples into $m$ independent problems, and hence the overall procedure has a cost of $O(m)$ per iteration. 
A similar approach, but with the minimization of $\ell_\infty$ norm of the re-projection error was proposed in \cite{mitra2008scalable}. 
It was shown to be more reliable and degraded gracefully with noise compared to $\ell_2$ based BA algorithms. 
Recently Wu proposed an incremental approach for bundle adjustment~\cite{wu2013towards}, 
where a partial BA or a full BA is performed after adding each camera and associated scene points to the set of unknown parameters, 
again with a complexity of $O(m)$. We use the ADMM framework to develop our approach. 

%

\subsection{Alternating Direction Method of Multipliers}
\label{sec:ADMM}
ADMM is a simple and powerful procedure well-suited for distributed optimization~\cite{lions1979splitting}, 
see also~\cite{boyd2011distributed}. 
In order to understand D-ADMM, consider the objective $h(x) := \sum_{i=1}^n h_i(x)$. 
We introduce local variables with a {\it consensus} equality constraint:
\begin{equation}
\label{eqn:ADMM_basic}
\begin{aligned}
\min_{\{x_i\},u} & \sum_{i=1}^n h_i(x_i) \\
\text{subject to } & x_i-u = 0, i \in \{1, \ldots, n\}.
\end{aligned} 
\end{equation}
To solve this problem, we first write down an {\it augmented Lagrangian}~\cite{rockafellar2009variational}:
\begin{align}
\label{eqn:ADMM_AL}
\overline l_\phi(x, u, r, \rho) := \sum_{i=1}^n h_i(x_i) + r_i^T(x_i-u) + \frac{\rho}{2} \phi(x_i,u),
\end{align} where $\rho>0$ is the penalty parameter, $r_i$ is the Lagrangian multiplier for the constraint, and $\phi(x_i,u)$ is the augmentation term that measures the distance individual variables $x_i$ and the consensus variable $u$. We then find a saddle point using three steps to update $\{x_i\}$, $u$, and $\{r_i\}$. 
Typically $\phi(x_i,u)$ is chosen to the squared Euclidean distance in which case \eqref{eqn:ADMM_AL} becomes the proximal Lagrangian~\cite{rockafellar2009variational}, but other distance or divergence measures can also be used.


\section{Algorithmic formulation}
\label{sec:approach}


\subsection{Distributed estimation of scene points and camera parameters}
\label{sec:distr_world_camera_points}
We distribute the estimation among both the scene points and the camera parameters as illustrated in Figure~\ref{fig:WP_CP_dist}. 
We estimate the camera parameter and the scene point corresponding to each image point independently, 
and then impose appropriate equality constraints. Eqn. (\ref{eqn:BA}) can be written as
\begin{gather}
\label{eq:full_obj}
\min_{\{x_i^j\}, \{y_j^i\}, \{x_i\}, \{y_j\}}  \sum_{j=1}^m \sum_{i \in S(j)} \phi_m(z_{i,j} -f(x_i^j,y_j^i)) ,\\
\label{eqn:DBA_WCP}
\text{such that } x_i^j = x_i, \text{ } \forall i, \text{ and } \{j : i \in S(j)\},\\
y_j^i = y_j, \text{ } \forall j, \text{ and } \{i \in S(j)\}.
\end{gather}The augmented Lagrangian, with dual variables $r_i^{j}$ and $s_j^{i}$, is given by 
\begin{align}
\nonumber
\sum_{j=1}^m \sum_{i \in S(j)} & \phi_m(z_{i,j} -f(x_i^j,y_j^i)) + r_i^{jT} (x_i^j-x_i) +  s_j^{iT} (y_j^i-y_j)\\
 \label{eqn:DBA_WCP_ADMM}
&+(\rho_x/2) \phi_a(x_i^j-x_i)+(\rho_y/2) \phi_a(y_j^i-y_j)
\end{align}  Here $\phi_a$ measures the distance between the distributed world points and their consensus estimates, and distributed camera parameters and their consensus estimates. For $\phi_m$ we compare squared Euclidean and Huber losses, and $\phi_a$ is always the squared Euclidean loss.

The ADMM iteration is given by 
\begin{align}
\nonumber
&(x_i^{j (k+1)},  y_j^{i (k+1)}) := \argmin_{\{x_i^j\}, \{y_j^i\}}  \phi_m(z_{i,j} -f(x_i^j,y_j^i)) \\
\nonumber
&+ r_i^{j{(k)}T} (x_i^j-x_i^{(k)})+ s_j^{i{(k)}T} (y_j^i-y_j^{(k)})\\
\label{eqn:DBA_WCP_ADMM_1}
&+(\rho_x/2) \phi_a(x_i^j-x_i^{(k)})+(\rho_y/2) \phi_a(y_j^i-y^{(k)}_j),
\end{align}

\begin{align}
\label{eqn:DBA_WCP_ADMM_2}
x_i^{(k+1)} :=& \frac{1}{|j:i \in S(j)|} \sum_{j: i \in S(j)} \left( x_i^{j (k+1)} + (1/\rho_x) r_i^{j(k)}\right),\\
\label{eqn:DBA_WCP_ADMM_3}
y_j^{(k+1)} :=& \frac{1}{|i \in S(j)|} \sum_{i \in S(j)} \left( y_j^{i (k+1)} + (1/\rho_y) s_j^{i(k)}\right),\\
\label{eqn:DBA_WCP_ADMM_4}
r_i^{j(k+1)} &:= r_i^{j(k)} + \rho_x \left( x_i^{j (k+1)} - x_i^{(k+1)}\right),\\
 \label{eqn:DBA_WCP_ADMM_5}
 s_j^{i(k+1)} &:= s_j^{i(k)} + \rho_y \left( y_j^{i (k+1)} - y_j^{(k+1)}\right).
\end{align} The equation (\ref{eqn:DBA_WCP_ADMM_1}) has to be solved for all $j \in S(i), i \in \{1, \ldots, m\}$, and it can be trivially distributed across multiple processes. When $\phi_m$ is squared $\ell_2$ distance, $(\ref{eqn:DBA_WCP_ADMM_1})$ can be solved using the Gauss-Newton method \cite{nocedal99}, where we repeatedly linearize $f$ around the current solution and update $(x,y)$. When $\phi_m$ is the Huber loss, we use limited memory BFGS (L-BFGS) \cite{nocedal99} to update the distributed scene points. Upon convergence, we will obtain the consensus estimates $x_i$ and $y_j$ for all scene points and cameras.

\subsubsection{Convergence Analysis}
We show that under certain assumptions the proposed D-ADMM algorithm in Section \ref{sec:distr_world_camera_points} converges, 
using the non-convex and non-smooth framework developed by \cite{wang2015global}.

\noindent \textbf{Theorem 1} The D-ADMM algorithm proposed in Section \ref{sec:distr_world_camera_points} to the stationary point of the augmented Lagrangian in \ref{eqn:DBA_WCP_ADMM} when:
\begin{enumerate}
\item $f(.,.)$ is the perspective camera projection model,
\item $\phi_m$ is any convex, smooth loss function, and $\phi_a$ is the squared Euclidean loss.
\item $\rho_x$ and $\rho_y$ are sufficiently large.
\end{enumerate}

\noindent \textbf{Proof}
Let $d_{ij}$ be the stack of $\{x_i^j, y_j^i\}$, and ${\hat{d}} = [\hat{d}_{ij}]_{\forall i, \forall j}$. Similarly each pair of consensus variables are stacked as the vector $\hat{c}_{ij} = [x_i^T y_j^T]^T$, and ${\hat{c}} = [\hat{c}_{ij}]_{\forall i, \forall j}$. ${\hat{d}}$ and ${\hat{c}}$ are respectively equivalent to ${x}$ and $y$ in \cite{wang2015global}. We show that the five assumptions (A1-A5) of \cite[Thm. 1]{wang2015global} are satisfied.

\begin{enumerate}
\item Given our assumptions, the objective function in (\ref{eqn:DBA_WCP}) is coercive, i.e., it tends to $\infty$ as ${\hat{d}} \rightarrow \infty$ (A1).
\item The feasibility and sub-minimization path conditions are also satisfied since the constraint matrices are easily seen to be full rank (A2-A3).
\item Each additive part of the objective $\phi_m(z_{i,j} -f(x_i^j,y_j^i))$ is restricted prox-regular if $\phi_m$ is a smooth convex function and $f$ is the perspective camera model. The gradient will be steep when $\tilde{z}_3$ in~\eqref{eqn:persp_proj} is less than some $\epsilon > 0$ and 
$\phi_m(z_{i,j} -f(x_i^j,y_j^i))$ is prox-regular for $\epsilon >0$;
hence  A4 in \cite[Thm. 1]{wang2015global} holds.
\item Our objective with respect to the consensus variable is identically 0, which is trivially regular (A5).
\end{enumerate}

Since all the assumptions hold, the iterative algorithm in eqns. (\ref{eqn:DBA_WCP_ADMM_1})-(\ref{eqn:DBA_WCP_ADMM_5}) converges to a stationary point of the augmented Lagrangian for sufficiently large $\rho_x$ and $\rho_y$.

\subsubsection{Time Complexity}
Optimizing (\ref{eqn:DBA_WCP_ADMM_1}) takes $O(l)$ time for each round of updates, since (\ref{eqn:DBA_WCP_ADMM_1}) must be solved $l$ times, with each solve requiring constant time. The time complexity of the consensus steps for camera parameters and world points given by (\ref{eqn:DBA_WCP_ADMM_2}) and (\ref{eqn:DBA_WCP_ADMM_3}) are $O(m)$ and $O(n)$ respectively. For the Lagrangian parameter updates given by (\ref{eqn:DBA_WCP_ADMM_4}) and (\ref{eqn:DBA_WCP_ADMM_5}), the time complexity is $O(l)$. Hence the dominant time complexity of the proposed algorithm is $O(l)$ for each round. Since the algorithm can be trivially parallelized, the complexity can be brought down to $O(1)$ for each round, if we distribute all the observations to individual processors. 

\subsubsection{Communication Overhead}
\label{sec:comm_overhead}
Considering a sparse UAV network, assume that each world point is imaged by $d$ cameras. Each camera needs to maintain a copy of the consensus world points $x_i$. Therefore to update $x_i$ using (\ref{eqn:DBA_WCP_ADMM_2}), each camera needs to obtain $d-1$ individual estimates of $x_i^j$ and send its version of $x_i^j$ to $d-1$ other cameras. Values $r_i^j$  can be updated locally in each camera, given $x_i^j$, $x_i$ and previous versions of $r_i^j$ using (\ref{eqn:DBA_WCP_ADMM_4}). Hence, for each world point we have a communication overhead of $3(d-1)d$ floating points per iteration (each world point is a 3D vector). Hence for $n$ world points, the communication overhead is $3(d-1)dn$ floating points per iteration, where $d$ depends on the distance of the camera from the scene. 

\subsubsection{Generalized Distributed Estimation}
\label{sec:gen_dist_est}

The problem~(\ref{eqn:DBA_WCP_ADMM_1}) 
requires each processor to estimate $p+q > 2$ parameters from a single 2D observation.
To control the variability of individual estimates as the algorithm proceeds, we generalize the approach to use more than one observation and hence more than one scene point and camera vector
during each update step. This generalized step provides flexibility to adjust the number of 3D scene points and cameras based on computational capability of each thread in a CPU or a GPU. 
We solve 
\begin{align}
\nonumber
&(X_i^{j (k+1)},  Y_j^{i (k+1)}) := \argmin_{\{X_i^j\}, \{Y_j^i\}} \phi_m(Z_{i,j} -f(X_i^j,Y_j^i)) \\ 
\nonumber
&+ r_i^{j{(k)}T} (X_i^j-X_i^{(k)}) + s_j^{i{(k)}T} (Y_j^i-Y_j^{(k)})\\
\label{eqn:DBA_mWCP_ADMM_1}
&+(\rho_x/2) \phi_a(X_i^j-X_i^{(k)})+(\rho_y/2) \phi
_a(Y_j^i-Y^{(k)}_j),
\end{align}
where 
\begin{align}
\nonumber
X_i^{j(k+1)} &:= \left[ x_{i_1}^{j(k)}\ x_{i_2}^{j(k)}\ \ldots \ x_{i_{\pi}}^{j(k)} \right]^T, \\ 
Y_j^{i(k+1)} &:= \left[ y_{j_1}^{i(k)}\ y_{j_2}^{i(k)}\ \ldots \ y_{j_{\kappa}}^{i(k)} \right]^T
\end{align} 

\section{Experiments}
\label{sec:expt}
We perform several experiments with synthetic data and real data to show the convergence of the re-projection error and the parameter estimates. We also compare the performance of the proposed approach a the centralized BA algorithm that we implemented using LM.  
The LM stops when the re-projection error drops below $10^{-14}$, or when the regularization parameter becomes greater than $10^{16}$. 
We implement our distributed approach in a single multi-core computer and not in a sparse UAV network, 
but our architecture is well-suited for a networked UAV application.


\subsection{Synthetic Data}
\label{sec:syn_data}
We simulate a realistic scenario, with smooth camera pose transition, and noise parameters consistent with real-world sensor errors.
Using the simulation, we evaluate the error in the estimated 3D scene point cloud and the camera parameters, 
and investigate how estimation error of camera pose affects the final tie points triangulation.

\begin{figure}[thb]
\begin{center}
  \includegraphics[width=2.0in]{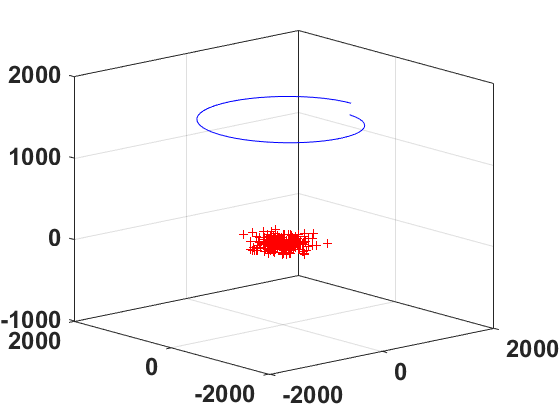}
\end{center}
   \caption{Camera flight path (blue) and 3D scene points (red) for an example synthetic data set.} 
\label{fig:syn_data}
\end{figure}

The camera positions are sampled around an orbit, with an average radius 1000m and altitude 1500m, with the camera  directed towards a specific area. To each camera pose, a random translation and rotation is added as any real observer cannot move in a perfect circle while steadily aiming always in the same exact direction. The camera path and the 3D scene points for an example scenario are shown in Figure \ref{fig:syn_data}. In practice, tie points are usually visible only within a small subset of the available views, and it is generally not practical to try to match all key points within each possible pair of frames. Instead, points are matched within adjacent frames. In our synthetic data, we create artificial occlusions or mis-detection so that each point is only visible on a few consecutive frames. 
 
\subsection{Convergence and Runtime}
\label{sec:converge}
\begin{figure}[thb]
\begin{center}
\hspace{-.1in}
  \subfigure[Rep. errors: $\phi_m$ in~\eqref{eq:full_obj} is $\ell_2$]{
  \includegraphics[width=1.6in]{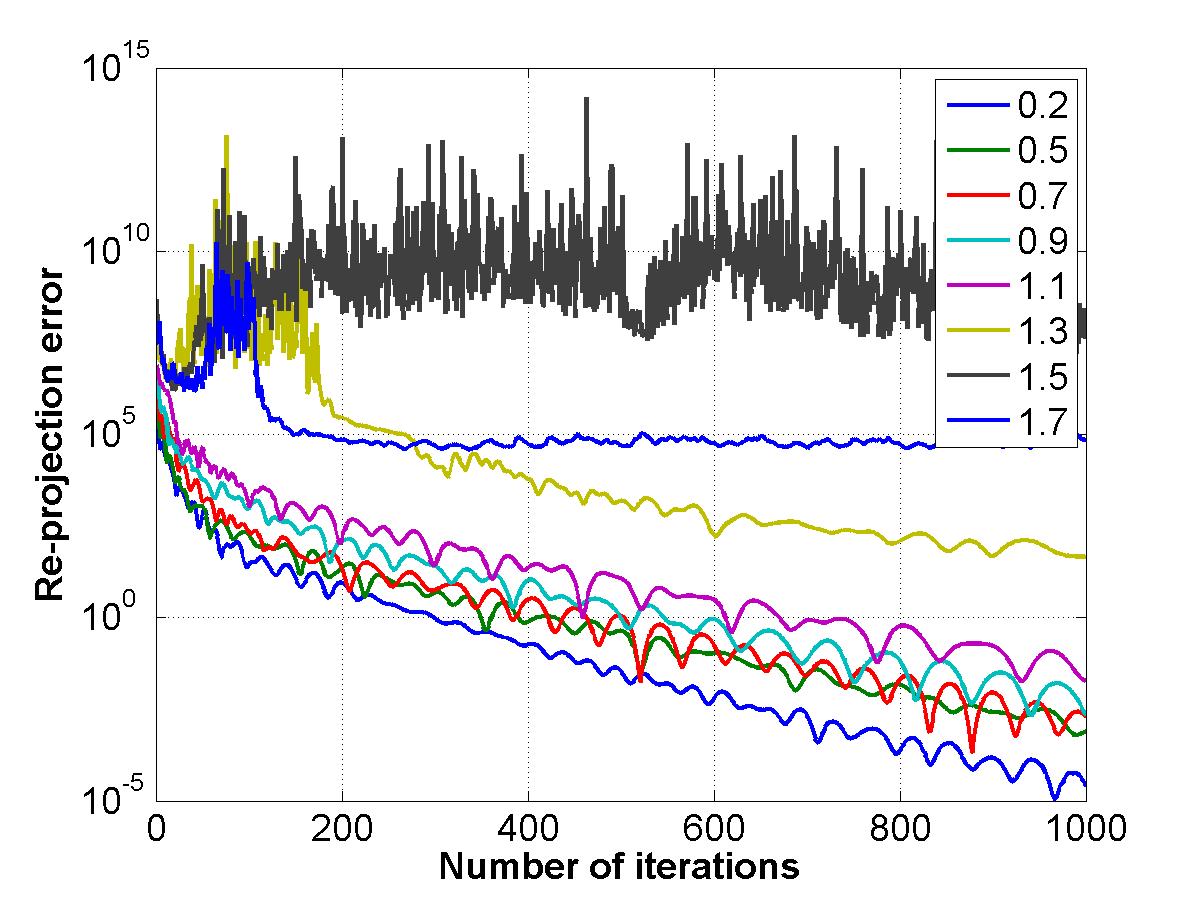}
  \label{LL_obj}}
\hspace{-.2in}
  \subfigure[MSE: $\phi_m$ in~\eqref{eq:full_obj} is $\ell_2$]{
  \includegraphics[width=1.6in]{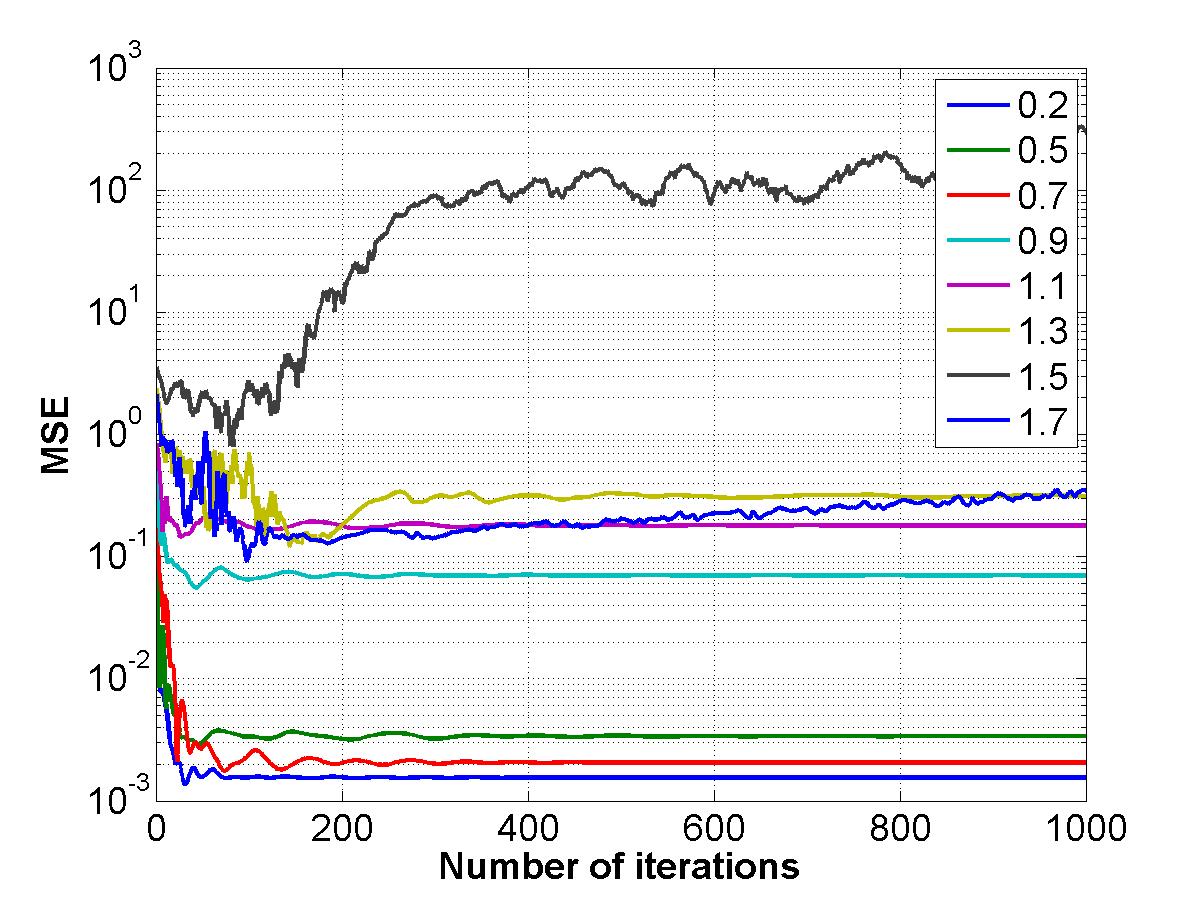}
\label{LH_obj}}
  \subfigure[Rep. errors: $\phi_m$ in~\eqref{eq:full_obj} is Huber]{
  \includegraphics[width=1.6in]{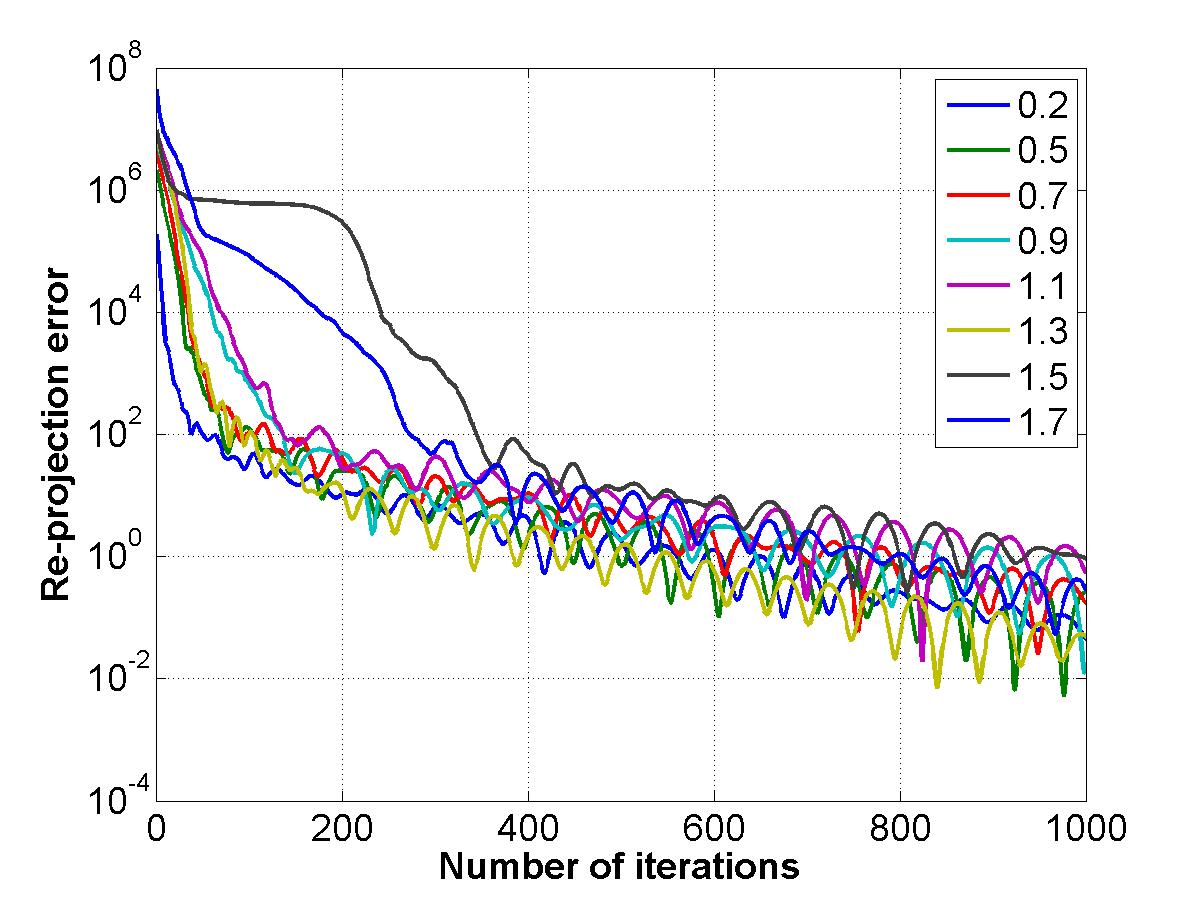}
  \label{HL_obj}}
  \hspace{-.2in}
   \subfigure[MSE: $\phi_m$ in~\eqref{eq:full_obj} is Huber]{
  \includegraphics[width=1.6in]{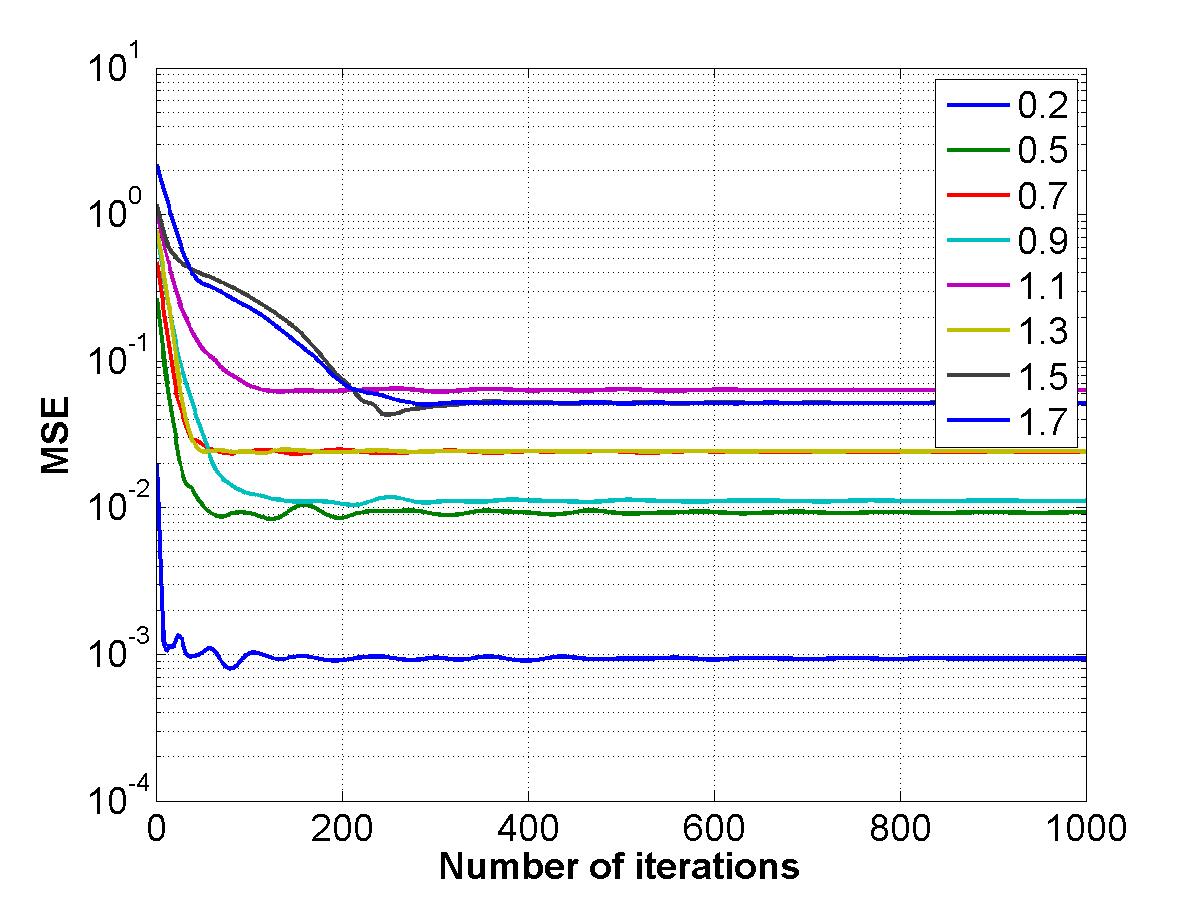}
  \label{HL_WP}}
\end{center}
   \caption{Choosing $\phi_m$ loss to be Huber penalty leads to better performance in distributed BA, 
   even when there are no outliers in the original data. Panels (a) and (c) compare reprojection errors, while 
   (b) and (d) compare MSE of scene points. In all figures,  curves correspond to values $\sigma$ of scene variance, as shown in the legend. Consensus penalty $\phi_a$ is always $\ell_2$.} 
\label{fig:obj_multi}
\end{figure}

We investigate convergence of the re-projection error and parameters for D-ADMM BA, comparing the convergence when $\phi_m$ is squared $\ell_2$ vs. Huber in~\eqref{eq:full_obj}, 
and $\phi_a$ always the squared $\ell_2$. The number of cameras is $5$, the number of scene points is $10$, and the number of 2D image points (observations) is $50$. We fix the standard deviation
for the additive Gaussian noise during the initialization of the camera angles and positions to be $0.1$. We vary the standard deviation of noise for the scene points from $0.2$ to $1.7$. Introducing robust losses for misfit penalty helps the convergence of the re-projection error significantly, see Figure~\ref{fig:obj_multi}, (a) vs. (c). This behavior is observed with the convergence of the scene points, see Figures~\ref{fig:obj_multi}, (b) vs. (d), and camera parameters. 
The Huber penalty is used to guard against outliers; here, outliers come from processors 
working with limited information. 
The performance degrades gracefully with noise, see Figure~\ref{fig:obj_multi}, (c) and (d).

\begin{figure*}[thb]
\begin{center}
  \subfigure[]{
  \includegraphics[width=2in]{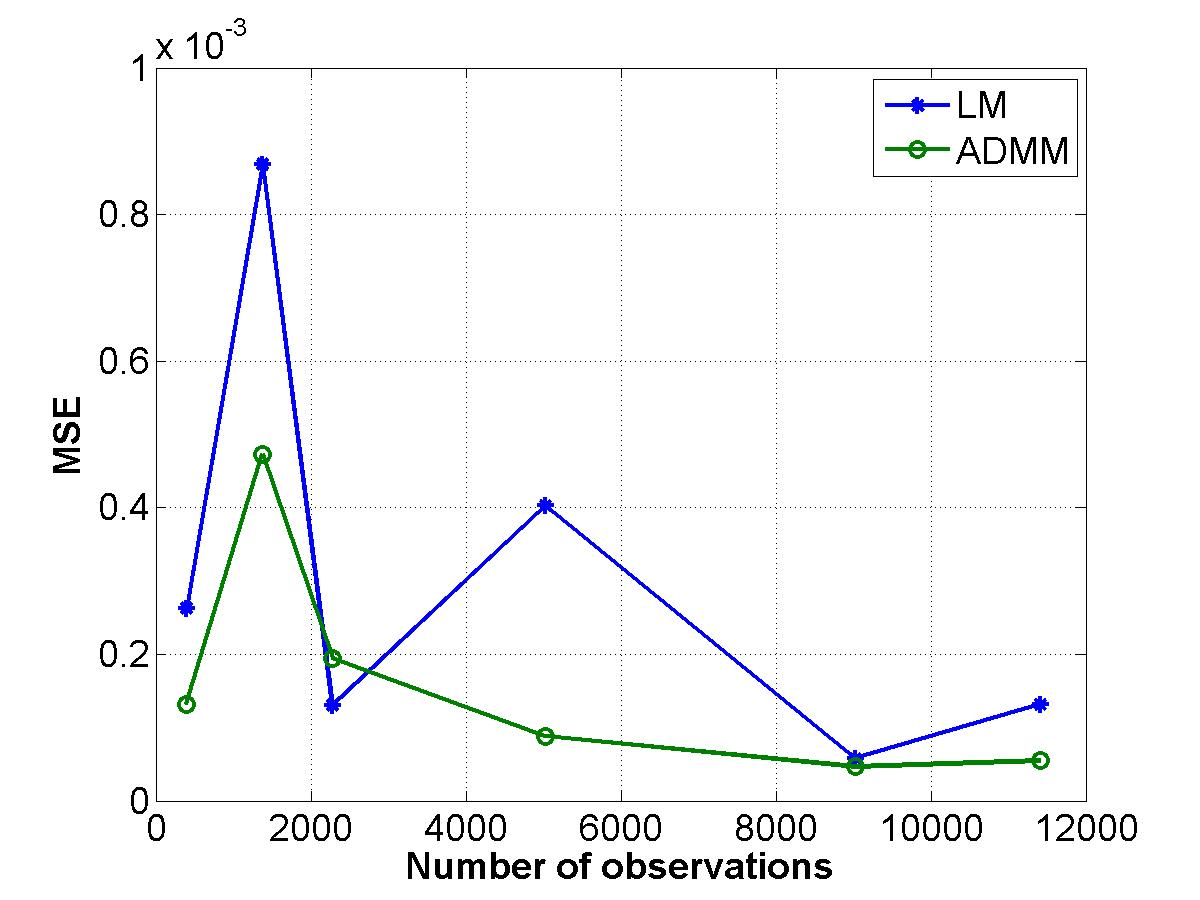}
  \label{CP_MSE}}
  \subfigure[]{
  \includegraphics[width=2in]{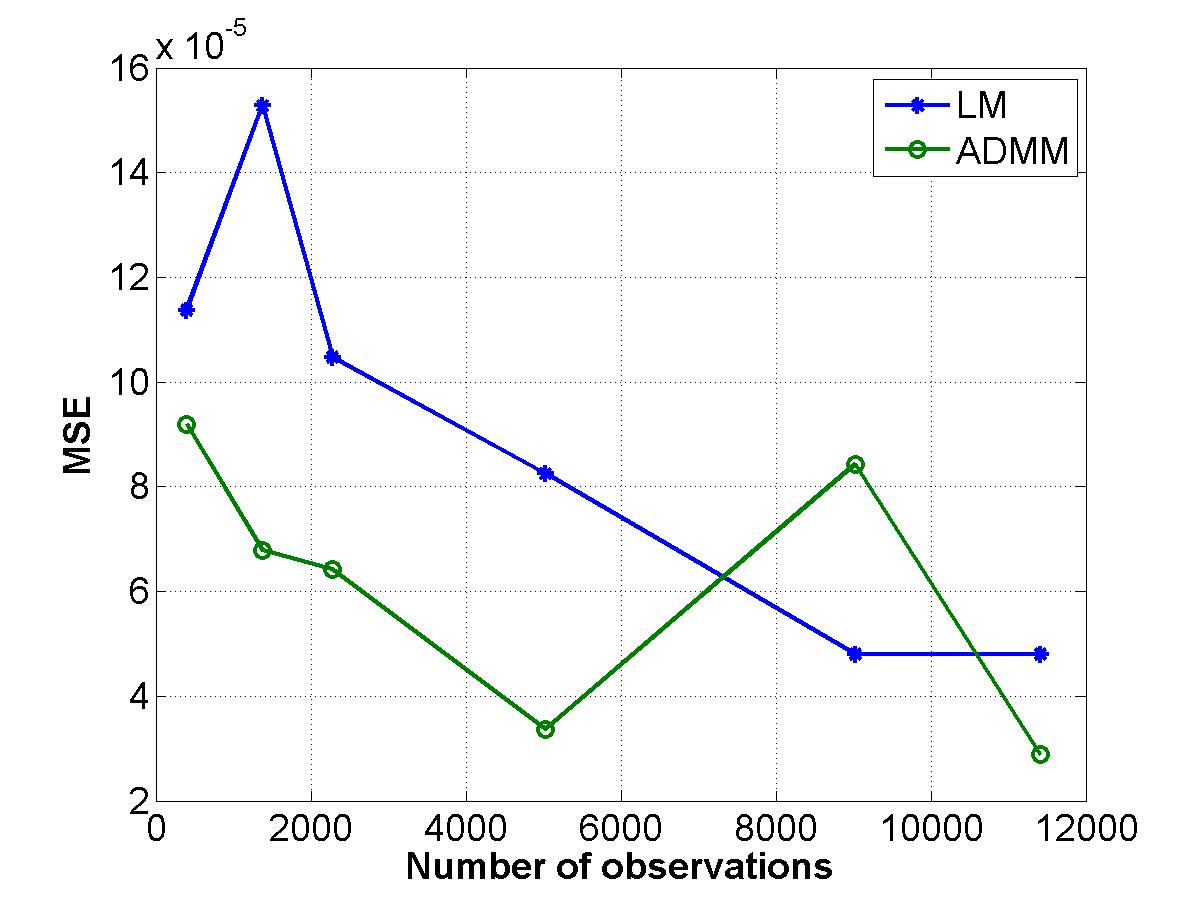}
  \label{WP_MSE}}
  \subfigure[]{
  \includegraphics[width=2in]{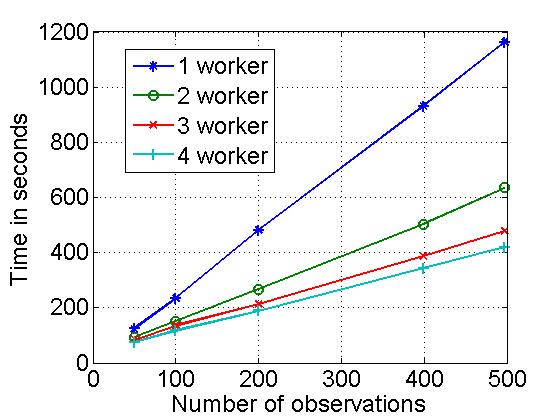}
  \label{Runtime}}
\end{center}
   \caption{(a) MSE between the actual and estimated camera parameters, (b) MSE between the actual and estimated scene points, (c) runtime of the proposed D-ADMM algorithm with increasing number of processor cores.} 
\label{fig:lm_vs_admm}
\end{figure*}

We also compare D-ADMM BA with the centralized LM BA and present the results in Figure \ref{fig:lm_vs_admm} (a) and (b). The number of camera parameters and 3D scene points are $(10, 40)$, $(15, 100)$, $(25, 100)$, $(30, 200)$, $(100, 200)$, and $(100, 250)$; with the number of observations increasing as shown on the x-axis of Figure~\ref{fig:lm_vs_admm}. In most settings, D-ADMM BA has a better parameter MSE than centralized LM BA. The runtime of the proposed approach with respect to the number of observations and parallel workers is shown in  Figure \ref{fig:lm_vs_admm} (c). The parallel workers are configured in MATLAB, and the runtime is linear with respect to the observations and reduces with increasing workers. Our implementation is a simple  demonstration of the capability of the algorithm --- a fully parallel implementation in a fast language such as C can realize its full potential.

\subsection{Real Data}
To demonstrate the performance of D-ADMM BA, we conducted experiments on real datasets with different settings. All experiments are done with MATLAB on a PC with a 2.7 GHz CPU and 16 GB RAM. 

In our SFM pipeline, SIFT feature points \cite{lowe1999object} are used for detection and matching. The relative fundamental matrices are estimated for each pair of images with sufficient corresponding points, which are used to estimate relative camera pose and 3D structure. Next, the relative parameters are used to generate the global initial values for BA.
The  datasets were downloaded from the Princeton Vision Group and the EPFL Computer Vision Lab \cite{strecha2008benchmarking}. 

\begin{figure*}[thb]
\begin{center}
  \subfigure[]{
  \includegraphics[width=1in]{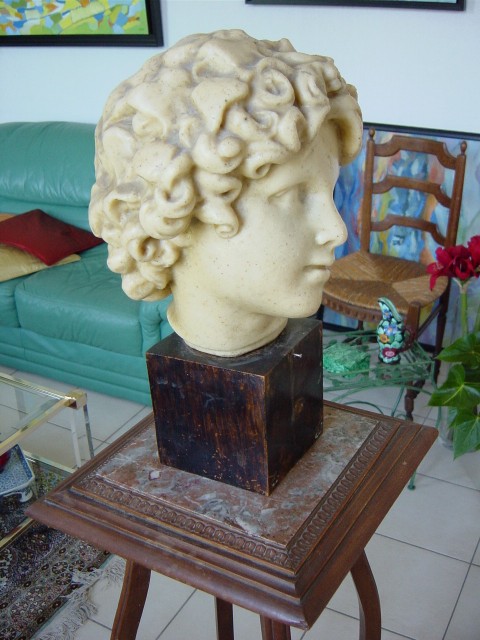}
  \label{B21}}
  \subfigure[]{
  \includegraphics[width=1.6in]{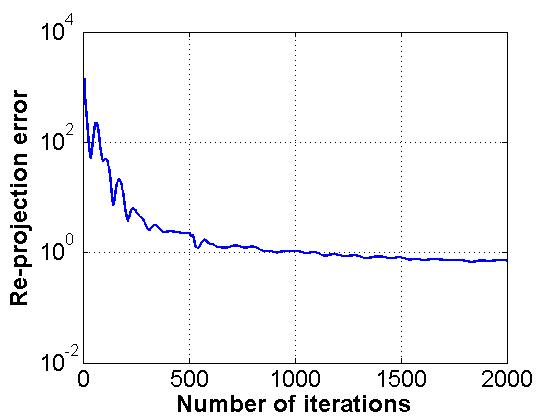}
  \label{obj_semilog_statue}}
  \subfigure[]{
  \includegraphics[width=1.4in]{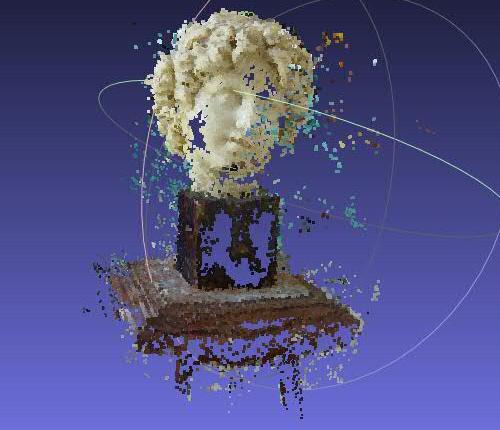}
  \label{dense1_lm}}
  \subfigure[]{
  \includegraphics[width=1.4in]{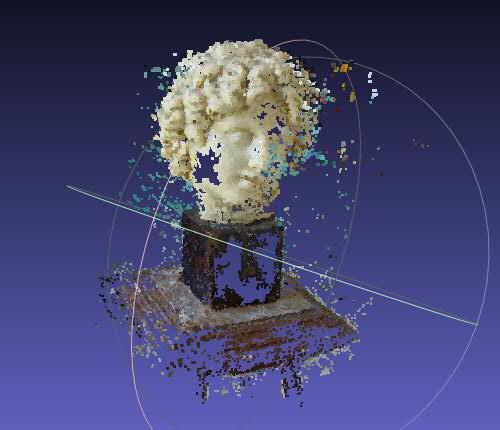}
  \label{dense1_admm}}
\end{center}
   \caption{(a) Original 2D image, (b) re-projection error for D-ADMM BA, (c) dense 3D point cloud estimated with LM BA (mean re-projection error = $0.93$), (d) dense 3D point cloud estimated using D-ADMM BA (mean re-projection error = $0.67$).} 
\label{fig:real}
\end{figure*}

\begin{figure}[thb]
\captionsetup[subfigure]{labelformat=empty}
\begin{center}
  \subfigure[]{
  \includegraphics[width=1.6in]{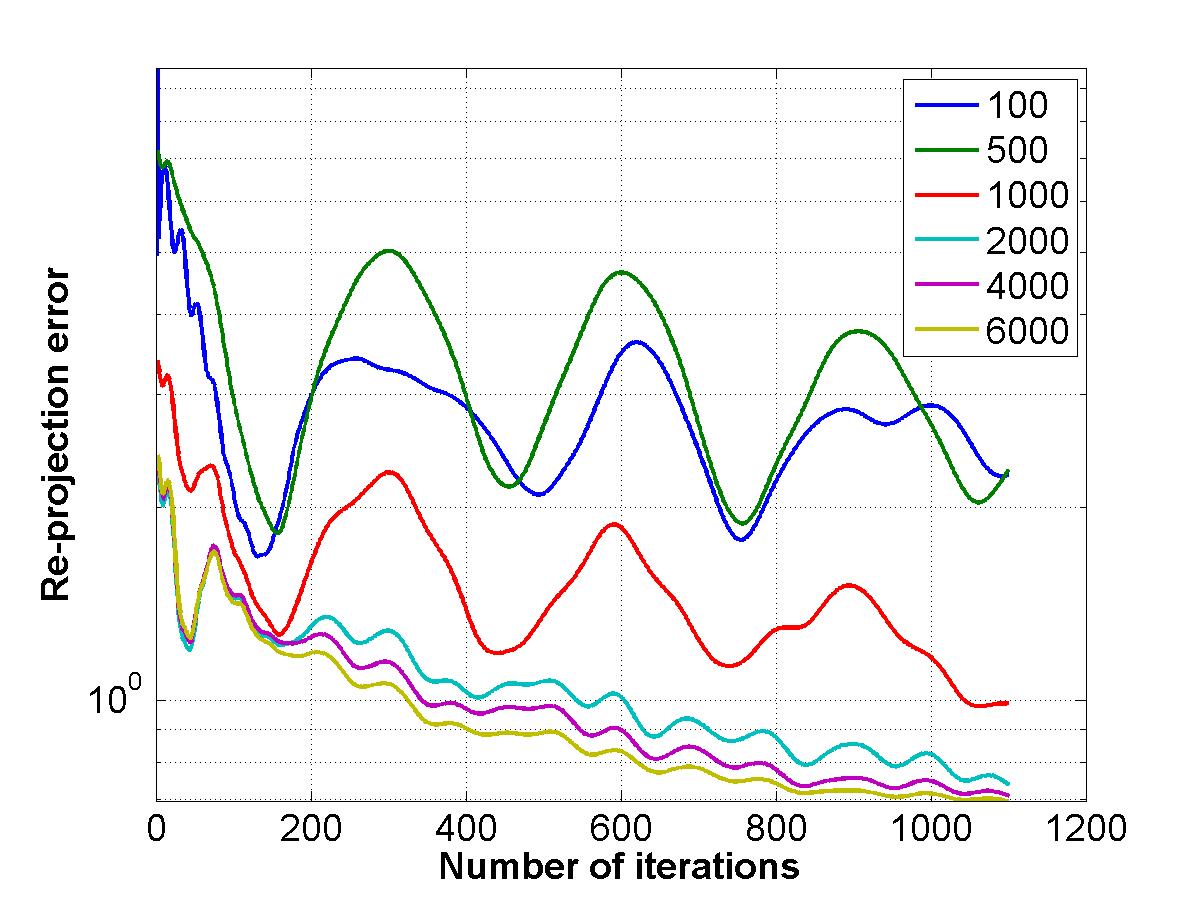}
  \label{im2_updated}}
  \hspace{-.2in}
  \subfigure[]{
  \includegraphics[width=1.6in]{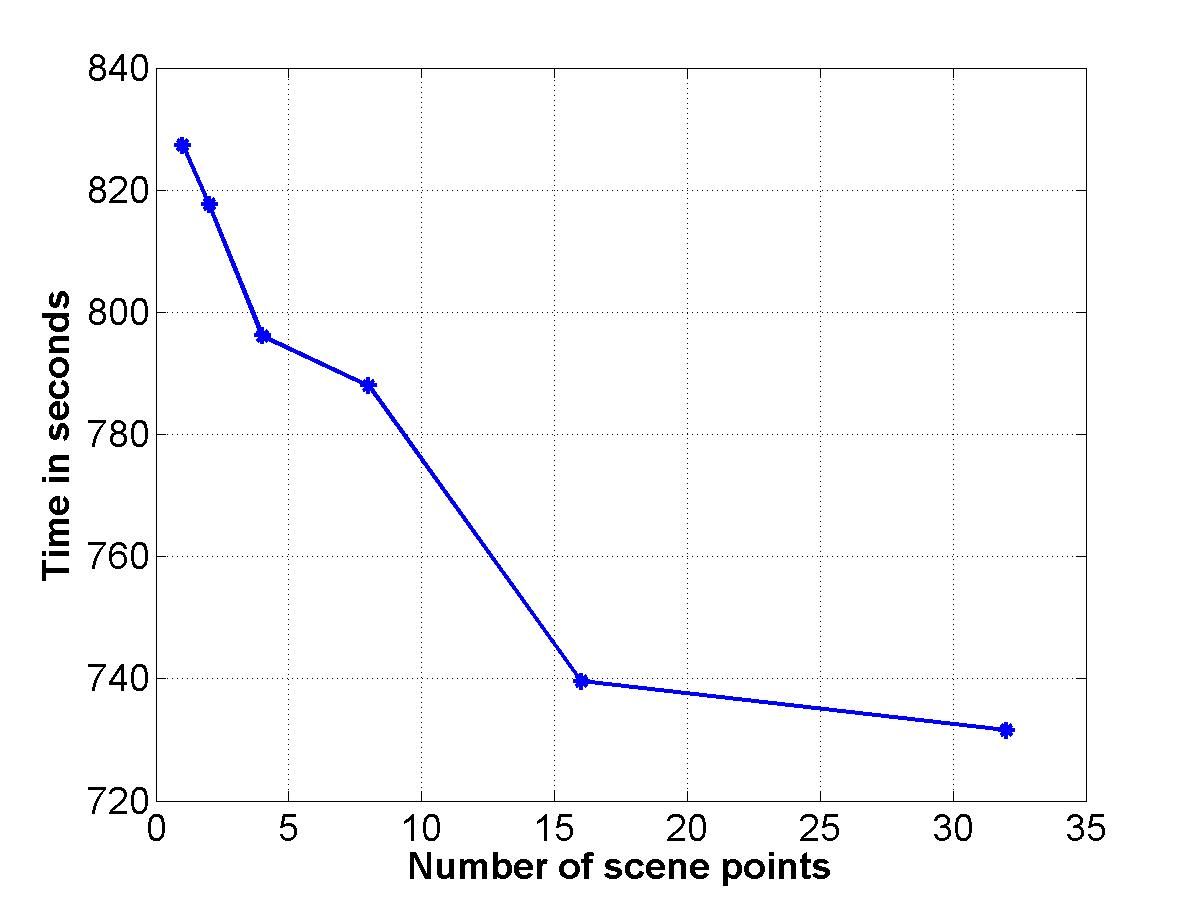}
  \label{im1_updated}}
  \subfigure[]{
  \includegraphics[width=1.6in]{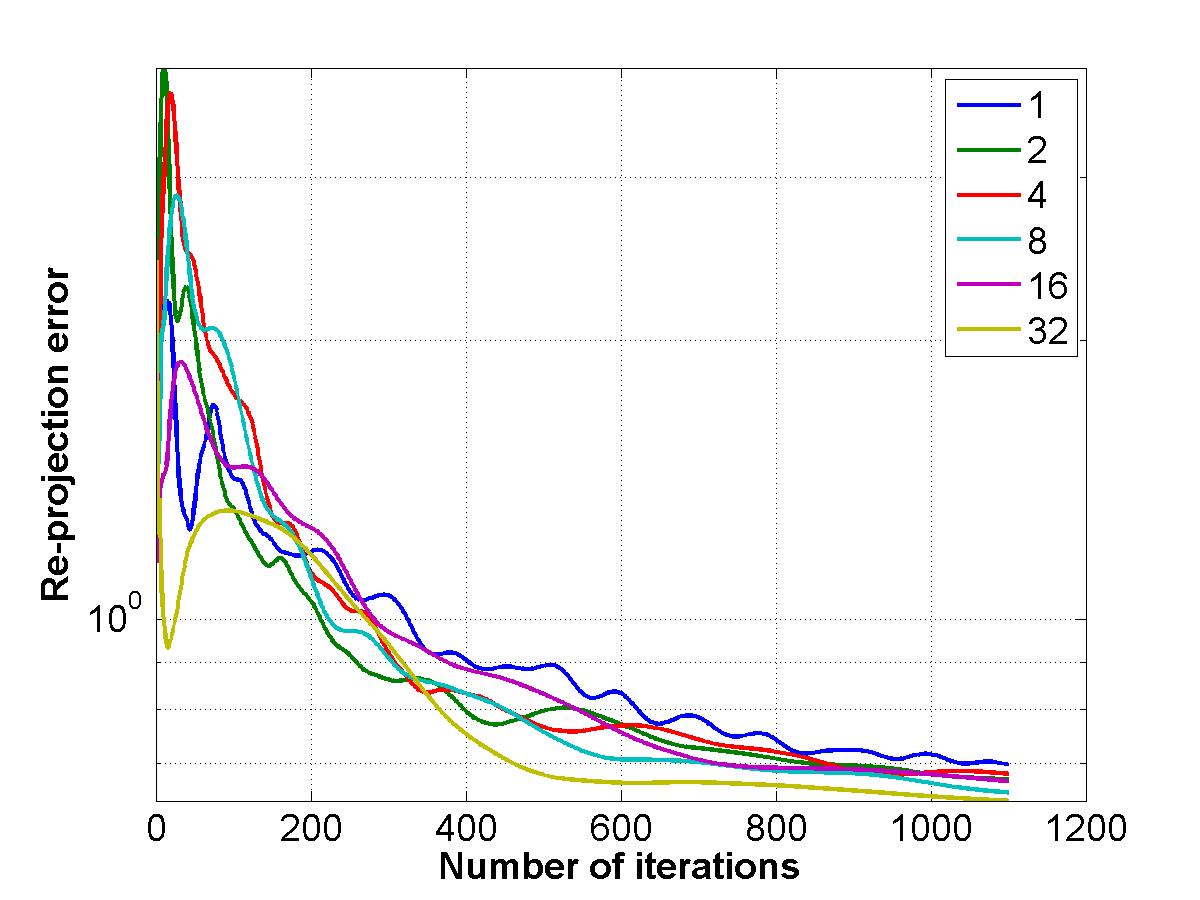}
  \label{im2_updated}}
    \hspace{-.2in}
  \subfigure[]{
  \includegraphics[width=1.6in]{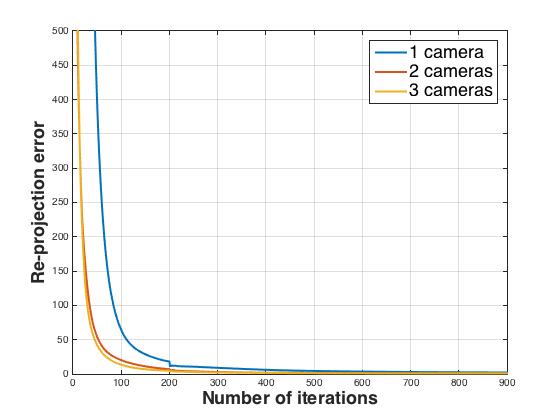}
  \label{compare_my}}
  \end{center}
   \caption{(a) The re-projection error for different values of $\rho$, using generalized distribution approach, (b) runtime of D-ADMM BA (c) re-projection errors with increasing number of scene points, (d) re-projection errors for multiple cameras per estimation vector.}
\label{fig:mr}
\end{figure}

Since there are no ground truth 3D structures available for the real datasets, we compare the dense reconstruction results obtained using the method of~\cite{xiao2008learning}. The first dataset has five images and a sample image is shown in Figure \ref{fig:real} (a). After keypoint detection and matching, centralized LM BA and D-ADMM BA are given the same input. There are a total of $104$ world points and $252$ observations. The final re-projection error of LM and D-ADMM are $0.93$ and $0.67$ respectively. 
Figure \ref{fig:real} (c) and (d) shows that the dense reconstruction quality of LM and the D-ADMM are similar. Figure \ref{fig:real} (b) shows the convergence of re-projection error for the D-ADMM algorithm. Figure \ref{fig:mr} (a) shows the convergence of re-projection error for different values of $\rho = \rho_x = \rho_y$. Setting $\rho$ to a high value accelerates  convergence. 

\begin{figure}[thb]
\begin{center}
  \subfigure[]{
  \includegraphics[width=1.25in]{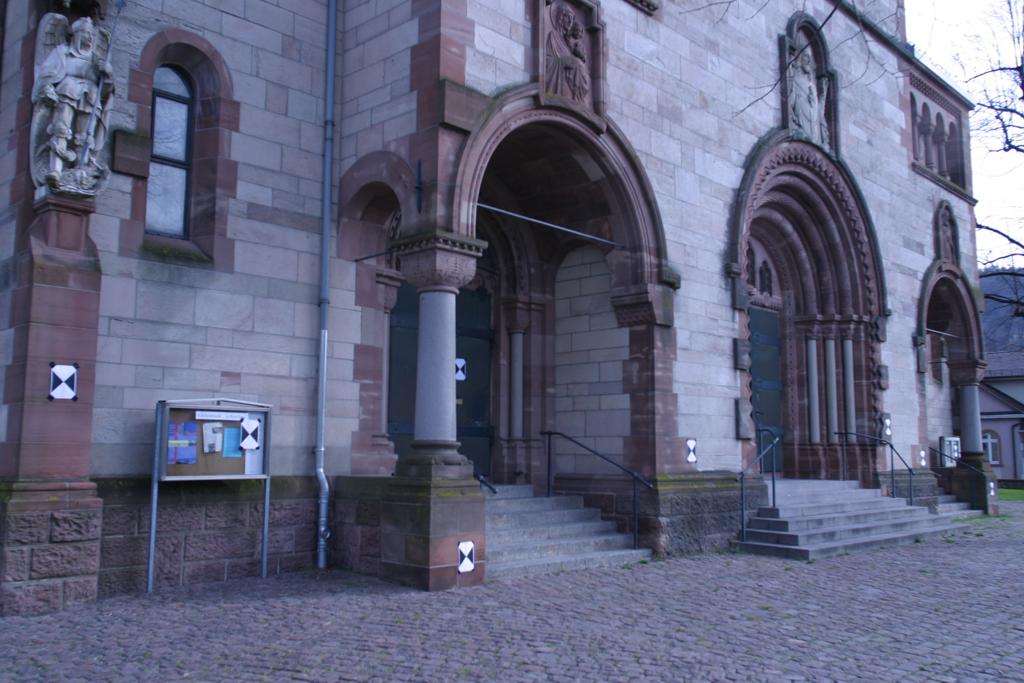}
  \label{orig_herzjesu}}
  \subfigure[]{
  \includegraphics[width=1.4in]{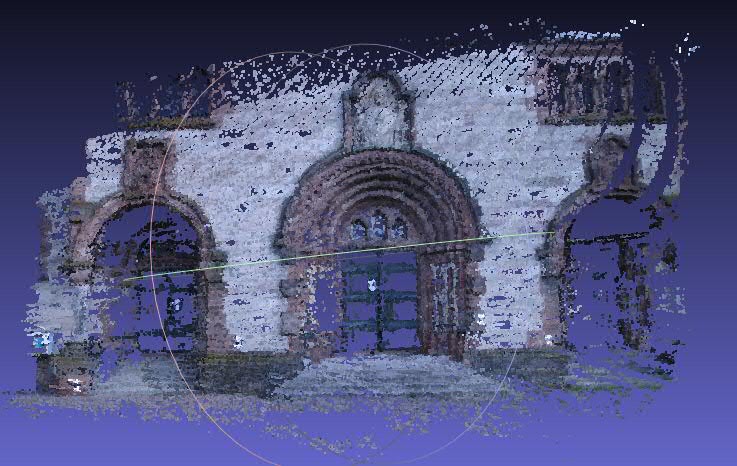}
  \label{dense1_admm_herzjesu}}
\end{center}
   \caption{(a) Original 2D image, (b) dense 3D point cloud estimated with D-ADMM BA (mean re-proj. error = $0.76$).} 
\label{fig:real_herzjesu}
\end{figure}

\begin{figure*}[thb]
\captionsetup[subfigure]{labelformat=empty}
\begin{center}
  \subfigure[]{
  \includegraphics[width=1.8in]{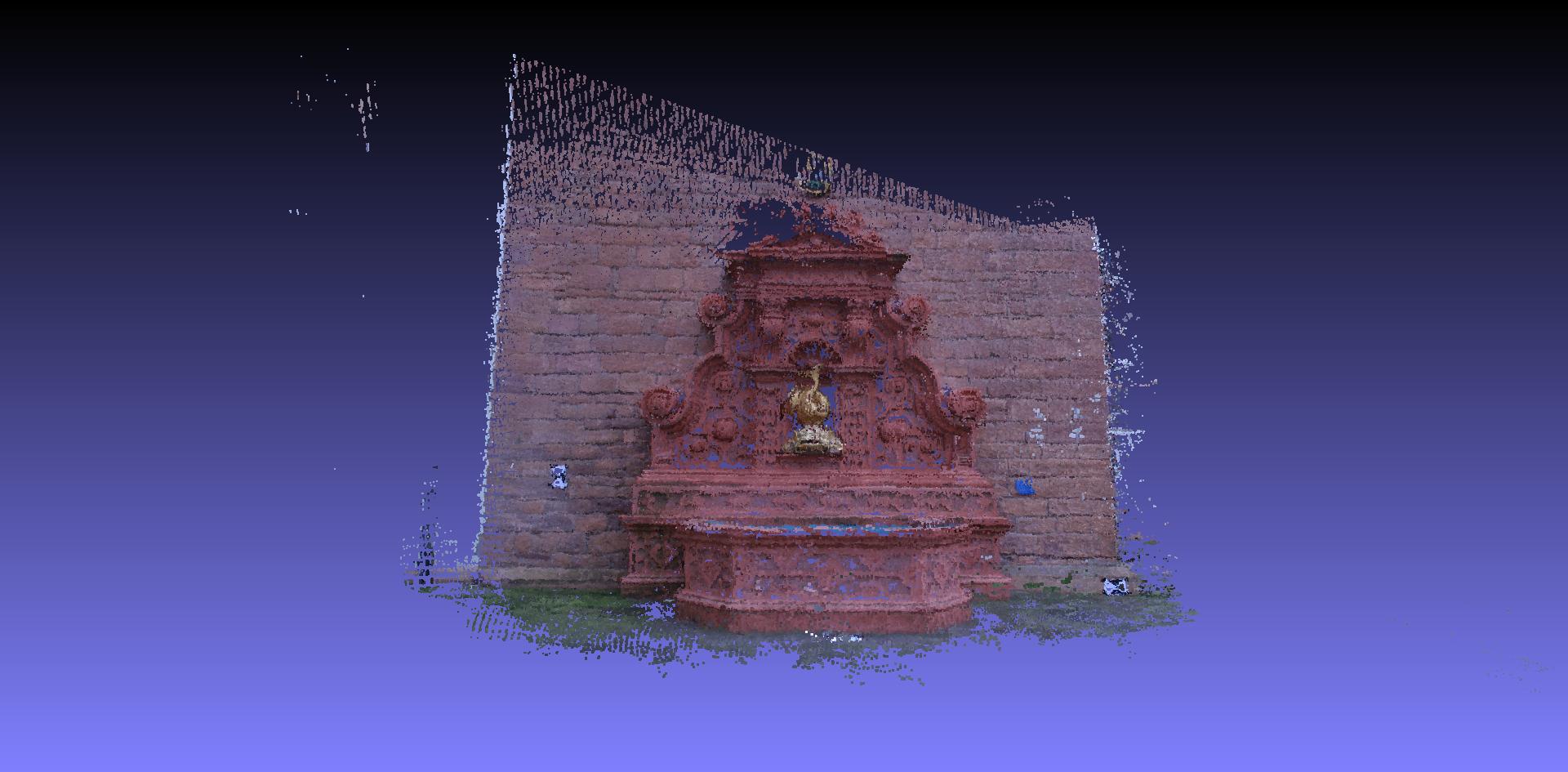}
  \label{im1_updated}}
  \subfigure[]{
  \includegraphics[width=1.8in]{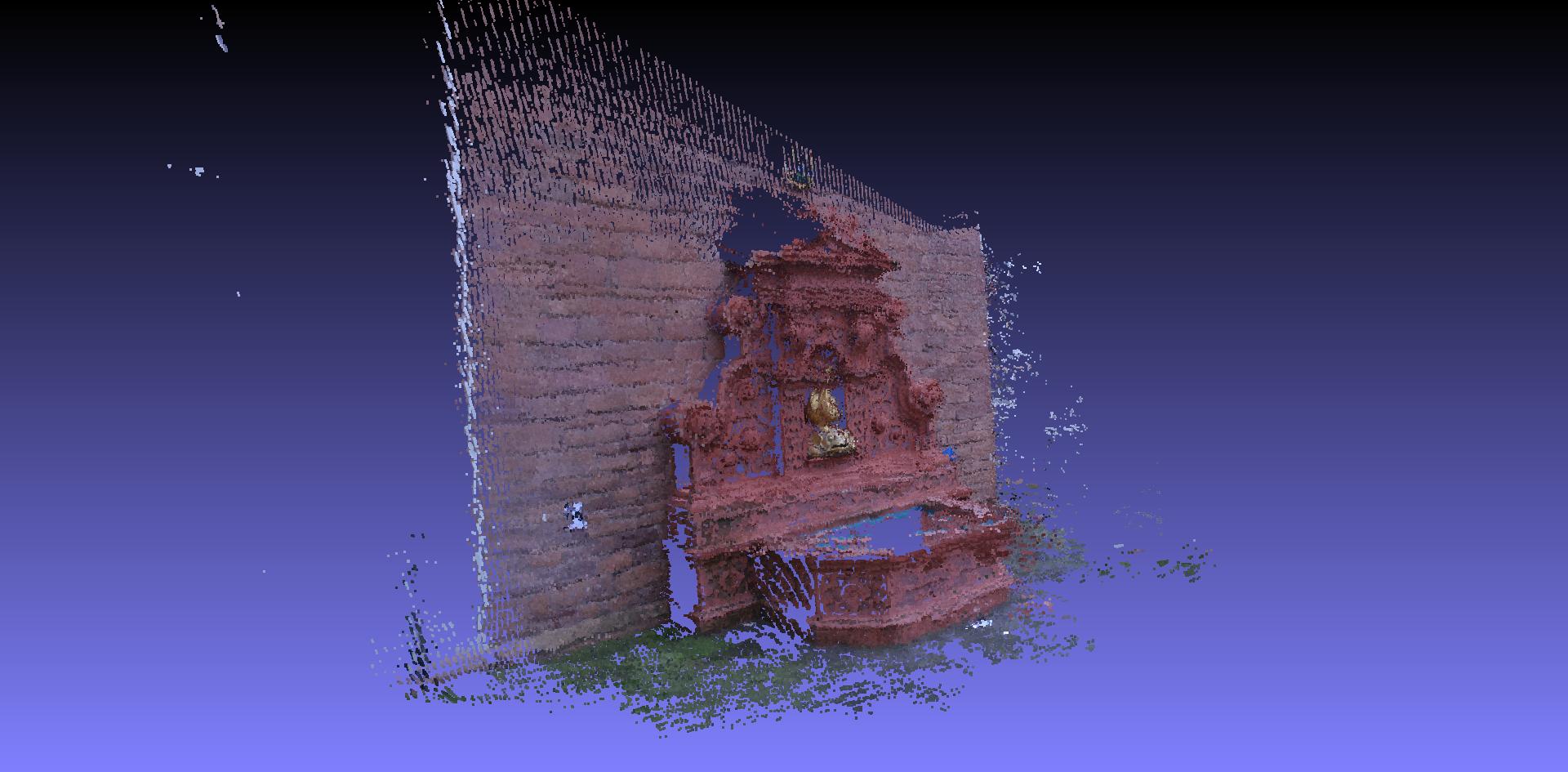}
  \label{im2_updated}}
  \subfigure[]{
  \includegraphics[width=1.8in]{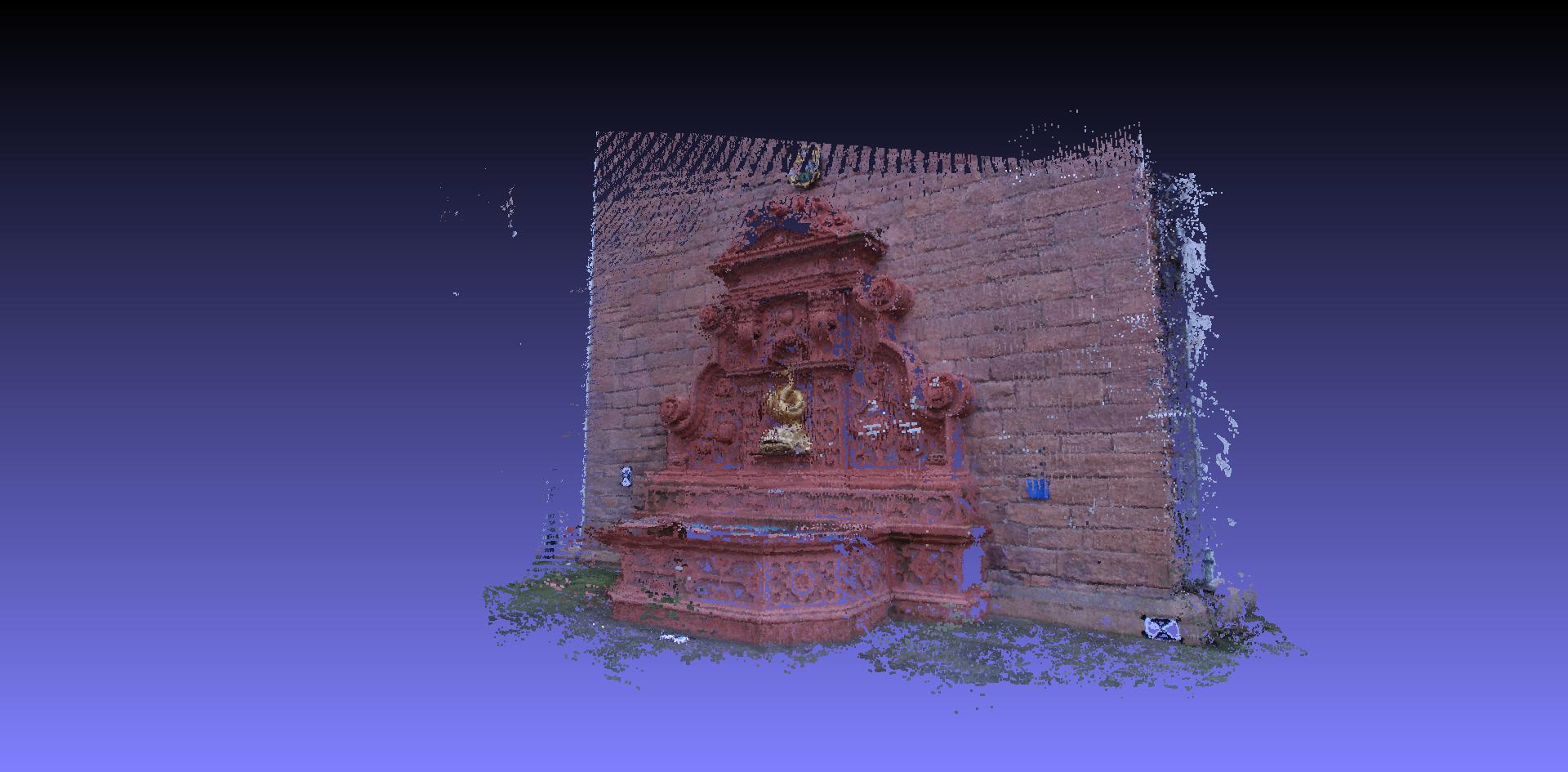}
  \label{im2_updated}}
\end{center}
   \caption{Reconstructed Fountain-P11 views (11 images, 1346 world points, 3859 obs., mean re-proj. error = $0.5$).}
\label{fig:fountain}
\end{figure*}

\begin{figure*}[thb]
\captionsetup[subfigure]{labelformat=empty}
\begin{center}
  \subfigure[]{
  \includegraphics[width=1.8in]{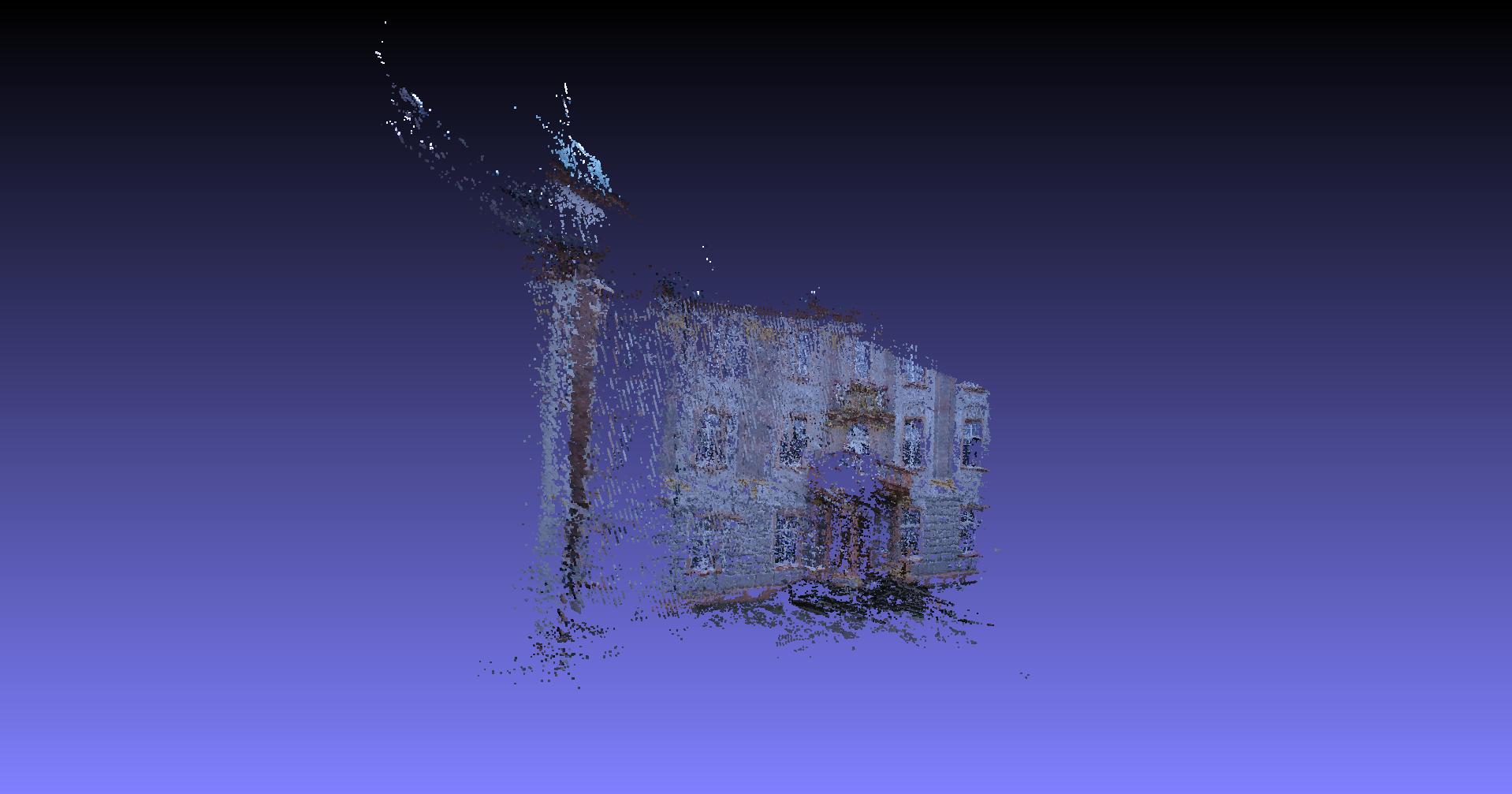}
  \label{im1_updated}}
  \subfigure[]{
  \includegraphics[width=1.8in]{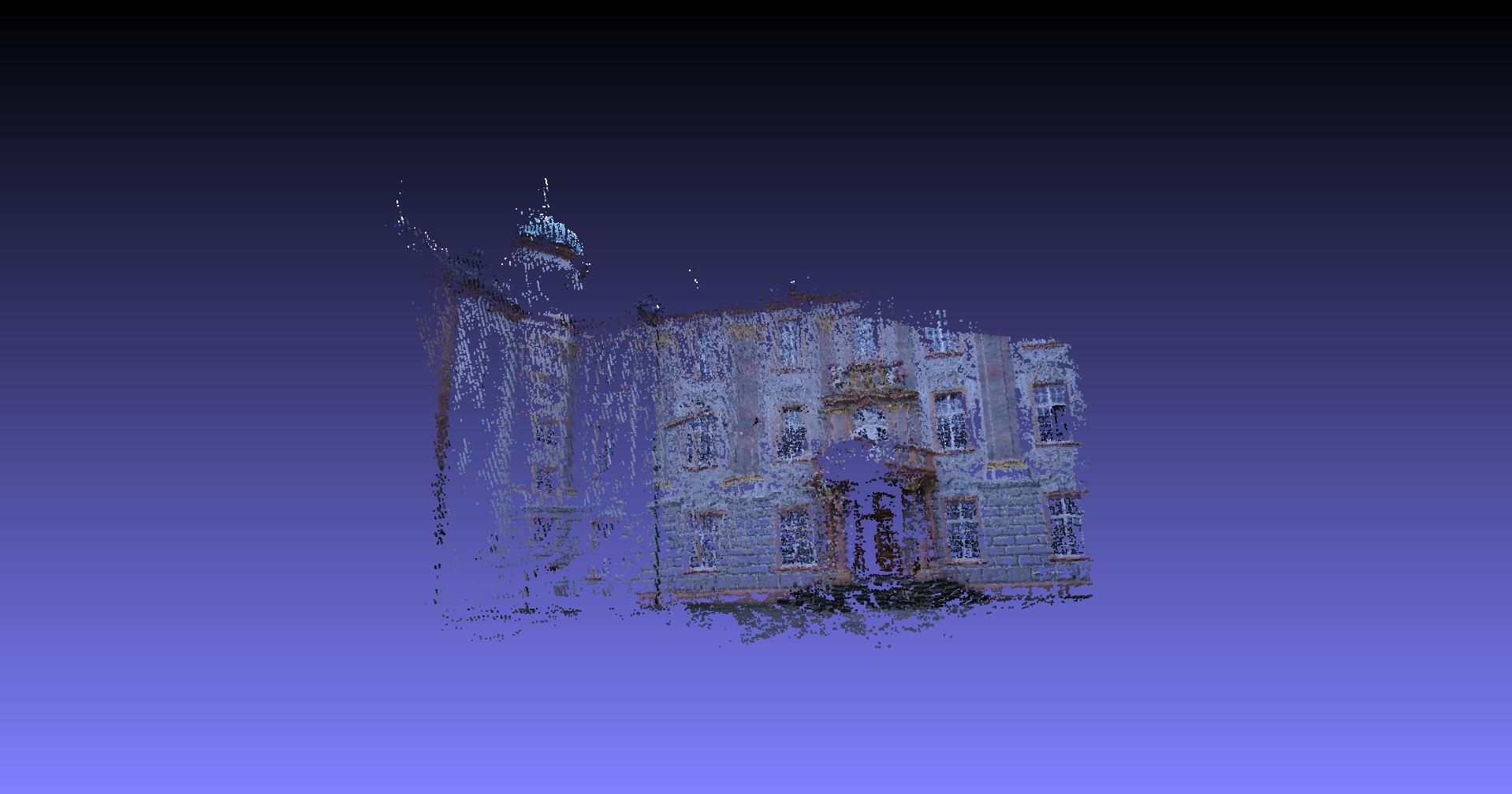}
  \label{im2_updated}}
  \subfigure[]{
  \includegraphics[width=1.8in]{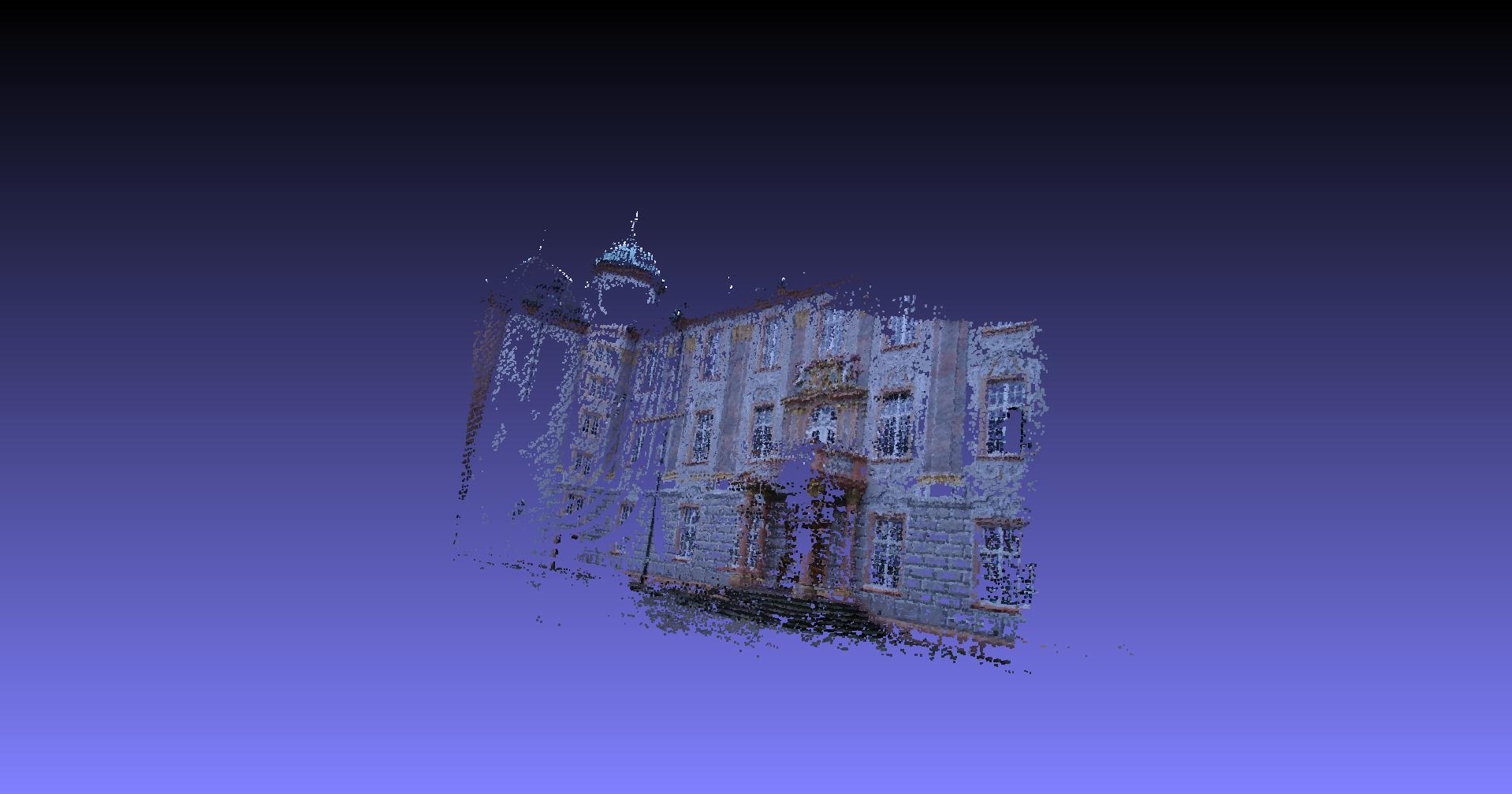}
  \label{im2_updated}}
\end{center}
   \caption{Reconstructed Entry-P10 views (10 images, 1382 world points, 3687 obs., mean re-proj. error = $0.7$).}
\label{fig:castle_entry}
\end{figure*}

\begin{figure*}[thb]
\captionsetup[subfigure]{labelformat=empty}
\begin{center}
  \subfigure[]{
  \includegraphics[width=1.8in]{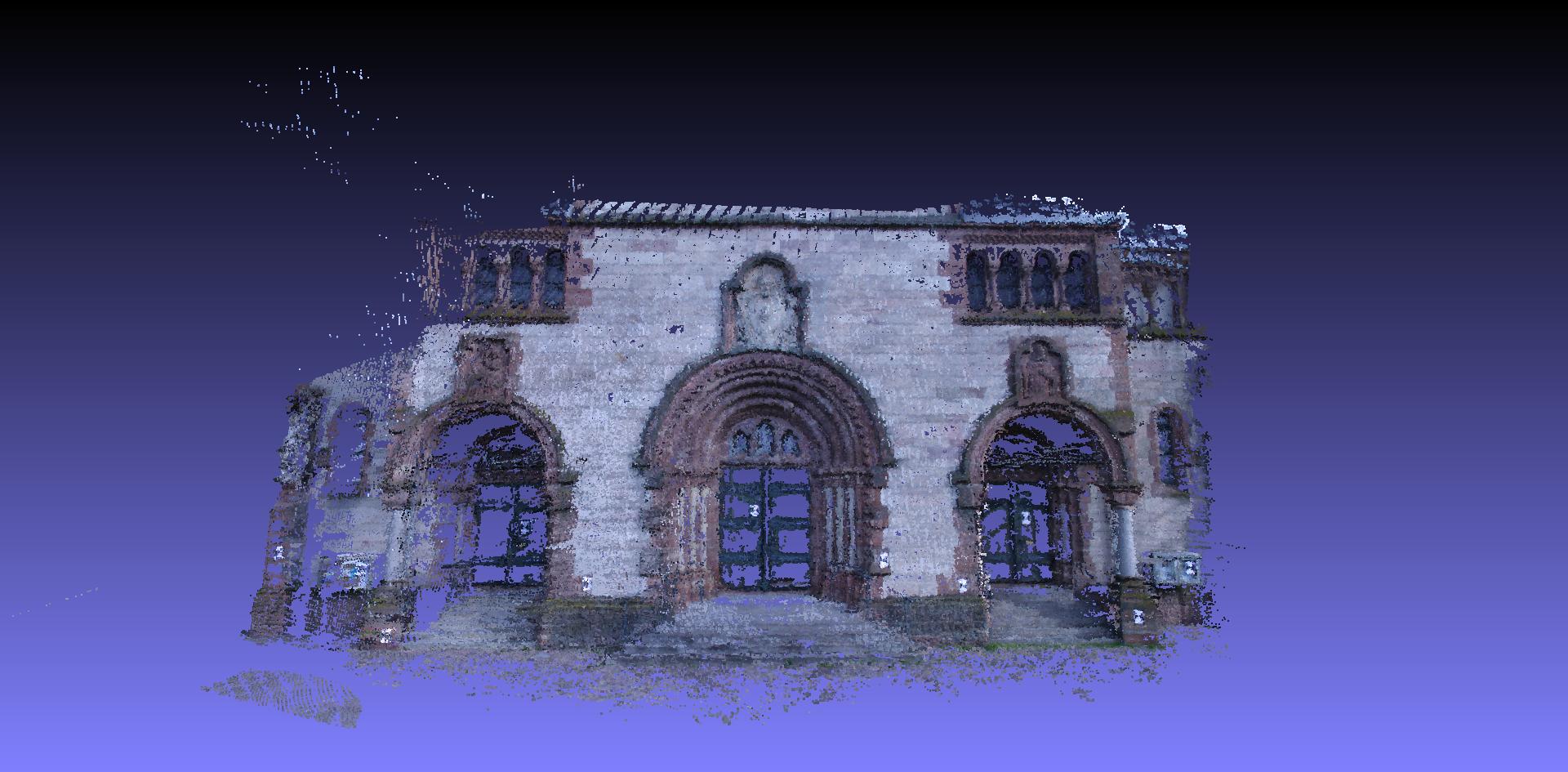}
  \label{im1_updated}}
  \subfigure[]{
  \includegraphics[width=1.8in]{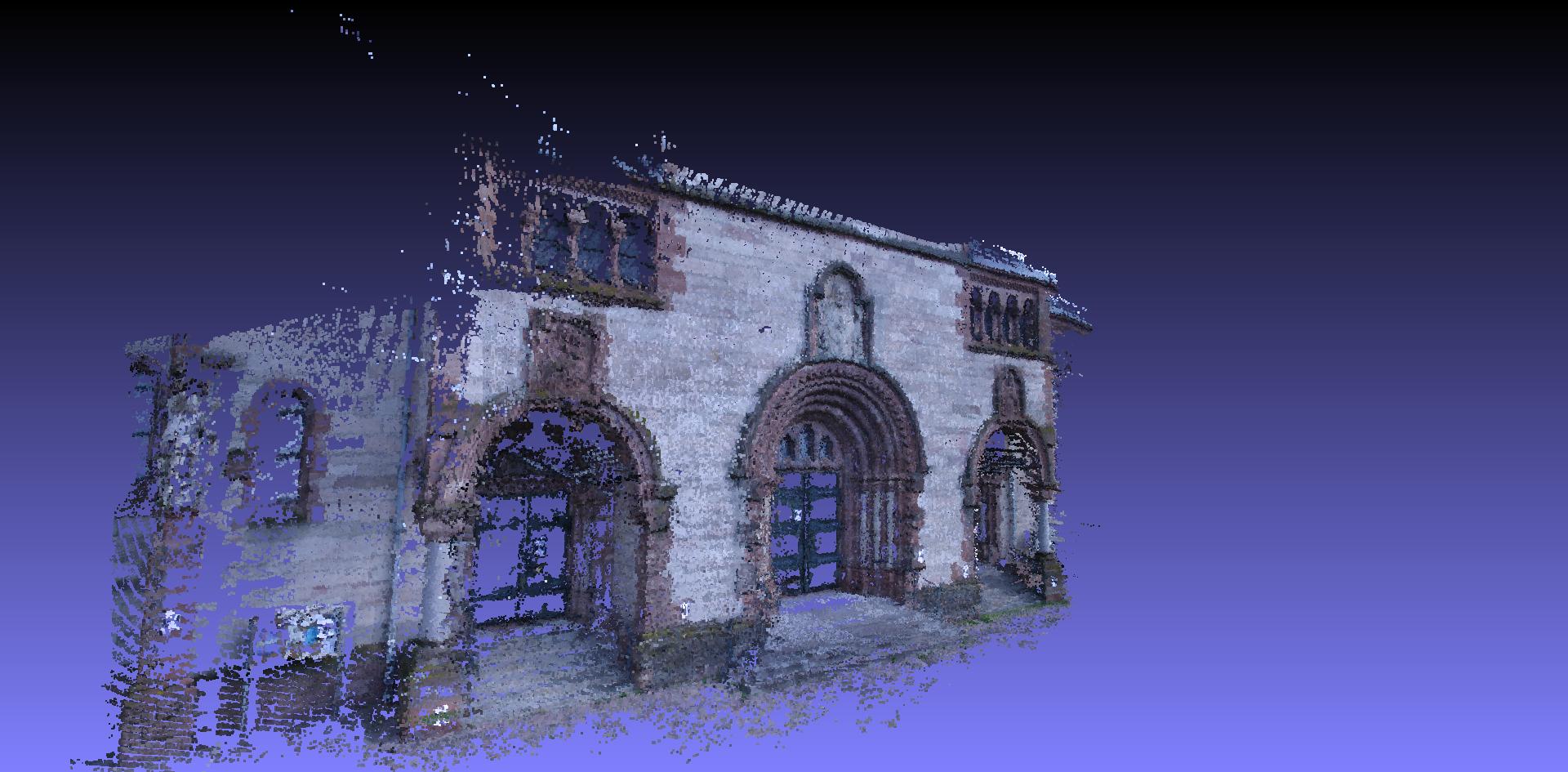}
  \label{im2_updated}}
  \subfigure[]{
  \includegraphics[width=1.8in]{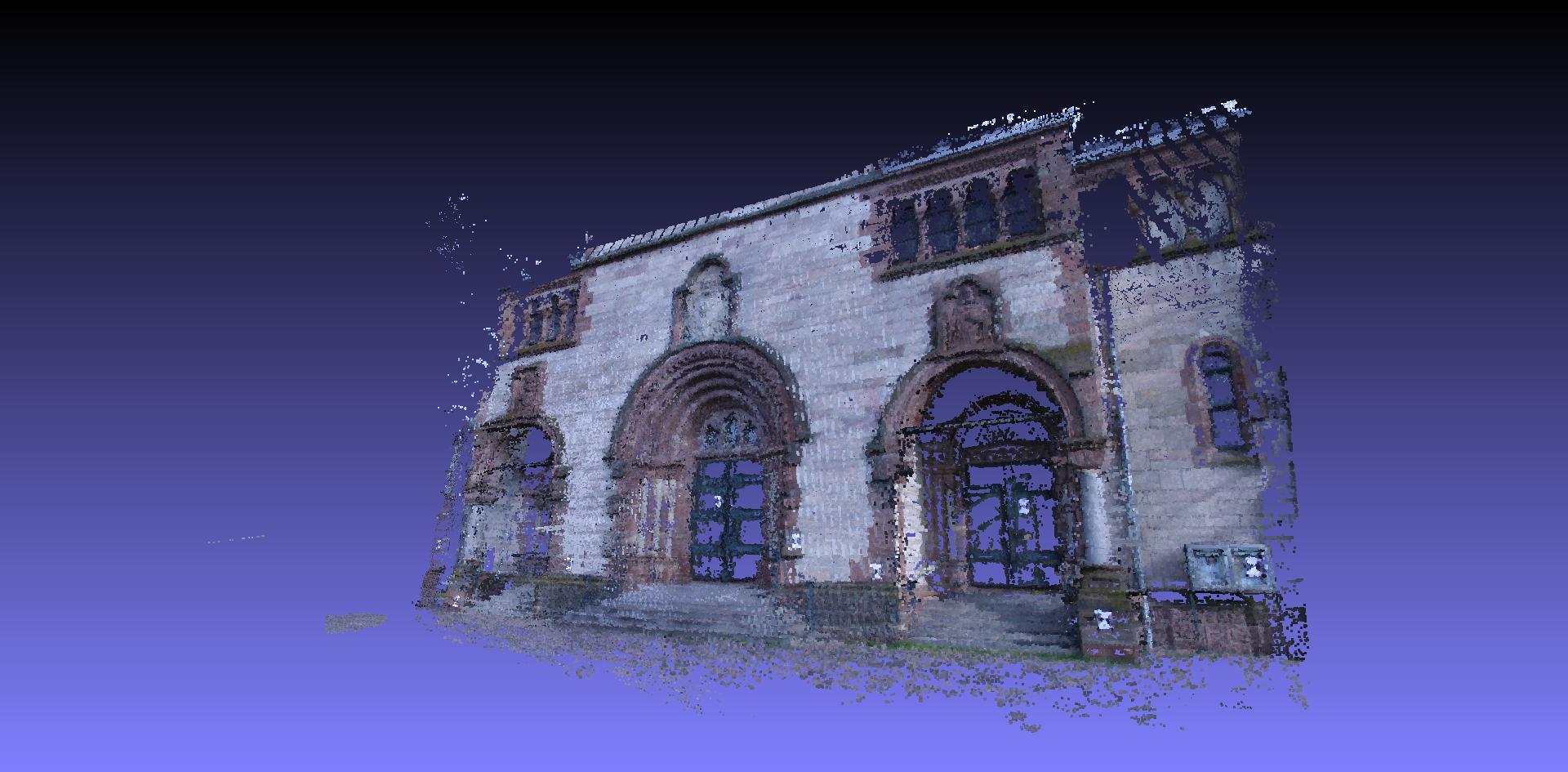}
  \label{im2_updated}}
\end{center}
   \caption{Reconstructed Herz-Jesu-P25 views (25 images, 2161 world points, 5571 obs., mean re-proj. error = $0.87$).}
\label{fig:herzjesu_large}
\end{figure*}

\begin{figure*}[thb]
\captionsetup[subfigure]{labelformat=empty}
\begin{center}
  \subfigure[]{
  \includegraphics[width=1.8in]{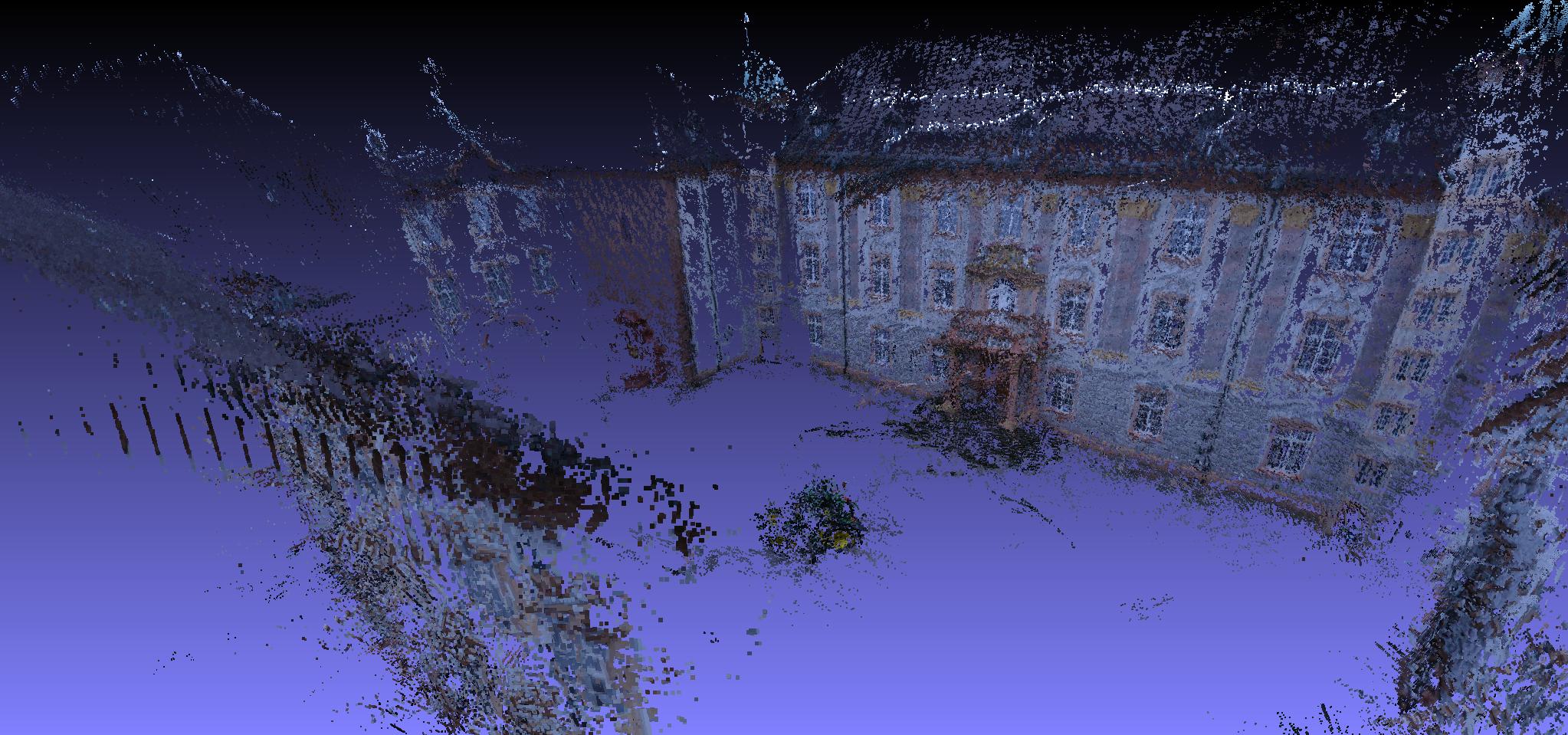}
  \label{im1_updated}}
  \subfigure[]{
  \includegraphics[width=1.8in]{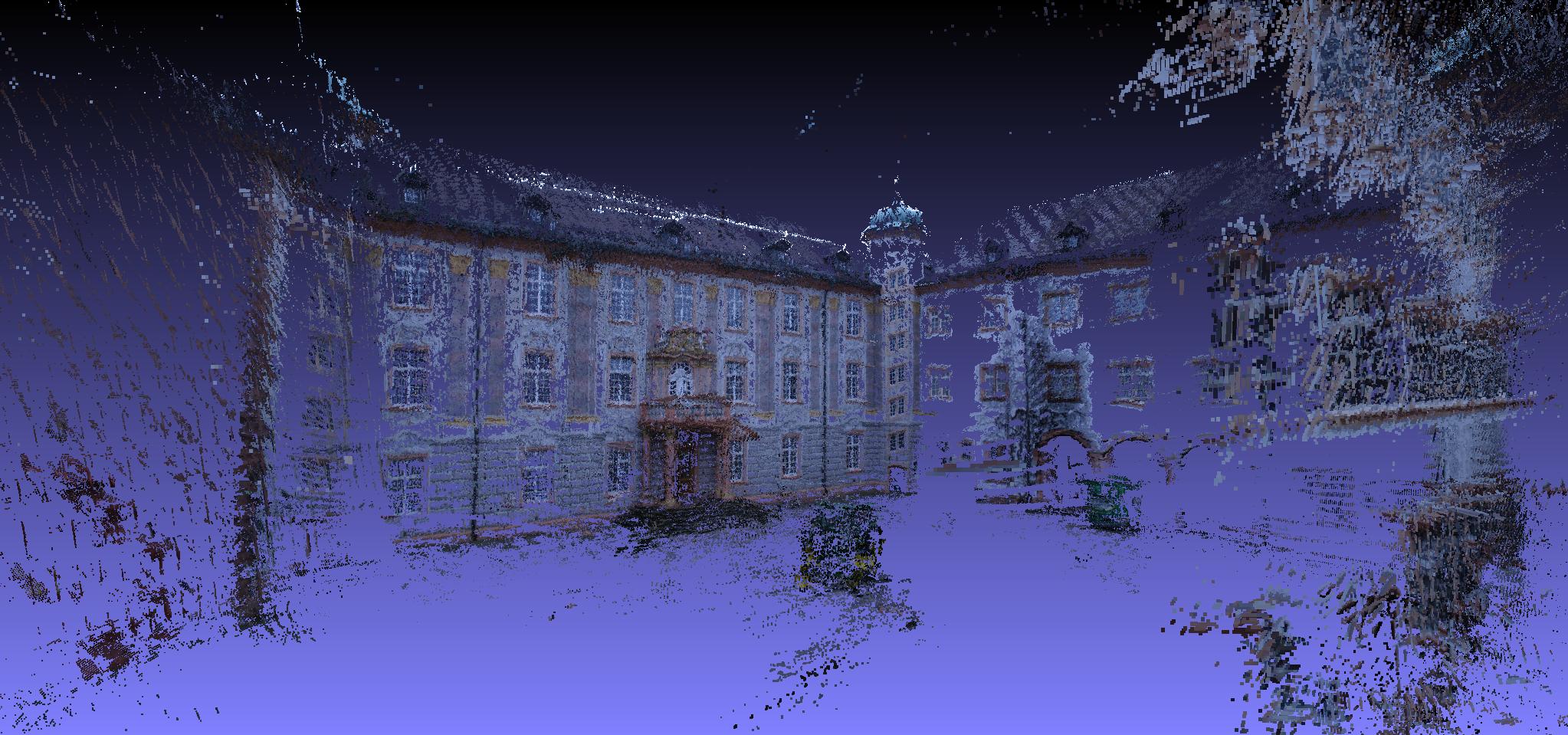}
  \label{im2_updated}}
  \subfigure[]{
  \includegraphics[width=1.8in]{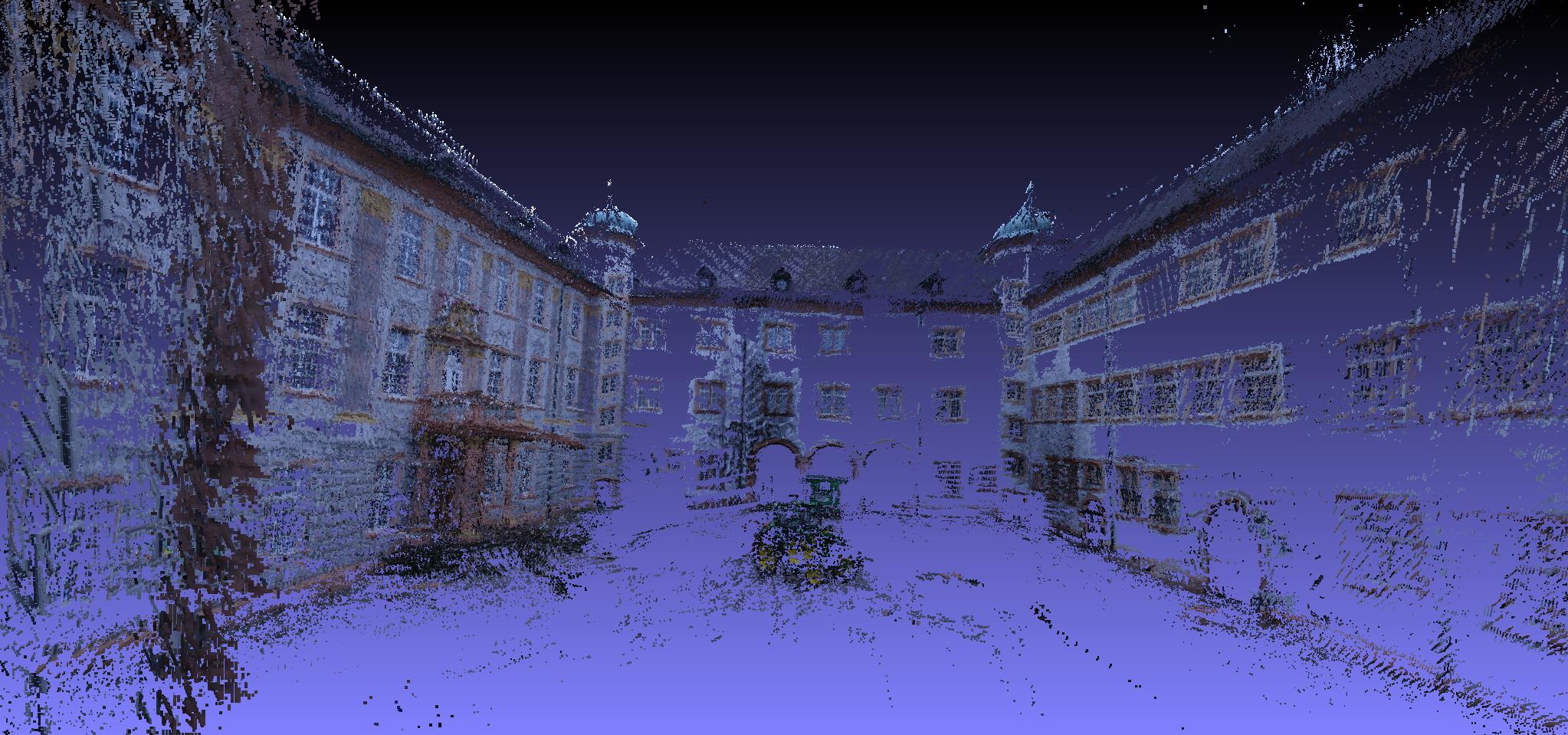}
  \label{im2_updated}}
\end{center}
   \caption{Castle-P30 (30 images, 2383 world points, 6453 obs., mean re-proj. error = $0.84$).}
\label{fig:castle_large}
\end{figure*}

We also estimate camera parameters and scene points, 
applying the approach of Section \ref{sec:gen_dist_est} to the same data set. Figure \ref{fig:mr} (b) shows that as the number of scene points per iteration increase, the runtime decreases, with $32$ scene points per iteration giving the fastest convergence, see~figure \ref{fig:mr} (c).  
Figure \ref{fig:mr} (d) compares re-projection errors with different number of cameras in each iteration. 
Initial values are the same as in the castle-P30 experiment (Figure \ref{fig:castle_large}), and the number of scene points in each iteration is $64$. Re-projection errors decrease faster as the number of cameras in each iteration increases.

We perform distributed BA on the Herz-Jesu dataset data set provided in  \cite{strecha2008benchmarking} using the  approach in Section \ref{sec:gen_dist_est}. This data set has seven images, 1140 world points, and 2993 observations. In this experiment, the LM BA algorithm using the same setting as in previous experiments does not converge and has the final re-projection error about $2500$. Therefore, the dense reconstruction result is not presented. D-ADMM BA with eight scene points in each update step has a final re-projection error of $0.76$. Figure \ref{fig:real_herzjesu}(b) shows the dense 3D point cloud estimated with D-ADMM BA. 

Additional results on other datasets (fountain-P11, entry-P10, Herz-Jesu-P25, and castle-P30) are presented in Table \ref{table:dataset_info}, Figure \ref{fig:fountain}, \ref{fig:castle_entry}, \ref{fig:herzjesu_large} and \ref{fig:castle_large}. 
$\sigma$ is mean re-projection error. Figure \ref{fig:fountain}, \ref{fig:castle_entry}, \ref{fig:herzjesu_large} and \ref{fig:castle_large} present different perspectives of the dense reconstruction results to show the robustness of 3D parameter estimations. 
\begin{table}[h!]
  \centering
  \caption{The dataset information and experiment results.}
  \begin{tabular}{ccccc}
    \toprule
    Dataset & Images & Scene pts & Obs & $\sigma$\\
    \midrule
    fountain-P11 & 11 & 1346 & 3859 & 0.5\\
    entry-P10 & 10 & 1382 & 3687 & 0.7\\
    Herz-Jesu-P25 & 25 & 2161 & 5571 & 0.87\\
    castle-P30 & 30 & 2383 & 6453 & 0.84\\
    \bottomrule
  \end{tabular}
  \label{table:dataset_info}
\end{table}

Settings are fixed across experiments, and the maximum iteration counter is set to $1600.$ The experiments on fountain-P11 and Herz-Jesu-P25 dataset (Figure \ref{fig:fountain} and \ref{fig:herzjesu_large}) have better dense reconstruction results since there are more images covering the same regions. 
 The real data experiments show D-ADMM BA achieves similar objective values (mean re-projection error $< 1$) as the number of observations increases; it is not necessary to increase the number of iterations as the size of the data increases. D-ADMM BA scales linearly with the number of observations and can be parallelized on GPU clusters.

\section{Conclusions}
\label{sec:concl}
We presented a new distribution algorithm for bundle adjustment, D-ADMM BA,  which compares well to centralized approaches in terms of performance and scales well for SfM. 
Experimental results demonstrated the importance of robust formulations for improved convergence in the distributed setting. 
Even when there are no outliers in the initial data, robust losses are helpful because estimates of 
processors working with limited information can stray far from the aggregate estimates, see Figure~\ref{fig:obj_multi}.
Formulation design for distributed optimization may yield further improvements; this is an interesting direction for future work. 

Results obtained with D-ADMM BA are
 comparable to those obtained with state-of-the-art centralized LM BA, 
 and D-ADMM BA scales linearly in runtime with respect to the number of observations.
Our approach is well-suited for use in a networked UAV system, 
where distributed computation is an essential requirement. 



\atColsBreak{\vskip5pt}
\FloatBarrier
\
{\small
\bibliographystyle{ieee}
\bibliography{ccbib}
}

\end{document}